\documentclass[letterpaper]{article} 
\usepackage{aaai25}  
\usepackage{times}  
\usepackage{helvet}  
\usepackage{courier}  
\usepackage[hyphens]{url}  
\usepackage{graphicx} 
\usepackage{natbib}  
\usepackage{caption} 
\usepackage{algorithm}
\usepackage{algorithmic}
\usepackage{multirow}
\usepackage{amsmath} 
\usepackage{subcaption}
\usepackage{booktabs}
\usepackage{newfloat}
\usepackage{listings}
\DeclareCaptionStyle{ruled}{labelfont=normalfont,labelsep=colon,strut=off} 
\floatstyle{ruled}
\newfloat{listing}{tb}{lst}{}
\floatname{listing}{Listing}
\usepackage{xcolor}
\newcommand{\revise}[1]{{{#1}}}
\title{Better Understandings and Configurations in MaxSAT Stochastic Local Search Solvers via Anytime Performance Analysis}
\author{
    Furong Ye\textsuperscript{\rm 1,2},
    Chuan Luo\textsuperscript{\rm 3}\thanks{Chuan Luo is the corresponding author of this work.},
    Shaowei Cai\textsuperscript{\rm 1}
}
\affiliations{
    \textsuperscript{\rm 1}Key Laboratory of System Software (Chinese Academy of Sciences) and State Key Laboratory of Computer Science, Institute of Software, Chinese Academy of Sciences, Beijing, China

    \textsuperscript{\rm 2} Leiden Institute of Advanced Computer Science (LIACS), Leiden University, Leiden, The Netherlands \\
    \textsuperscript{\rm 3} School of Software, Beihang University, Beijing, China \\

    f.ye@ios.ac.cn, chuanluo@buaa.edu.cn, caisw@ios.ac.cn
}

\usepackage{bibentry}

\begin{document}

\maketitle

\begin{abstract}
Though numerous solvers have been proposed for the MaxSAT problem, and the benchmark environment such as MaxSAT Evaluations provides a platform for the comparison of the state-of-the-art solvers, existing assessments were usually evaluated based on the quality, e.g., fitness, of the best-found solutions obtained within a given running time budget. However, concerning solely the final obtained solutions regarding specific time budgets may restrict us from comprehending the behavior of the solvers along the convergence process. This paper demonstrates that Empirical Cumulative Distribution Functions can be used to compare MaxSAT stochastic local search solvers' anytime performance across multiple problem instances and various time budgets. The assessment reveals distinctions in solvers' performance and displays that the (dis)advantages of solvers adjust along different running times. This work also exhibits that the quantitative and high variance assessment of anytime performance can guide machines, i.e., automatic configurators, to search for better parameter settings. Our experimental results show that the hyperparameter optimization tool, i.e., SMAC, can achieve better parameter settings of solvers when using the anytime performance as the cost function, compared to using the metrics based on the fitness of the best-found solutions.
\end{abstract}

\begin{links}
\link{Code and Datasets}{https://github.com/FurongYe/AAAI25-MaxSAT/}
\end{links}
\section{Introduction}
The Maximum Satisfiability (MaxSAT) problem is the optimization version of the influential Boolean Satisfiability (SAT) problem that aims at finding the maximum number of satisfied clauses~\cite{biere2009handbook}.
(Weighted) partial MaxSAT, which is an important generalization of MaxSAT, divides the clauses into hard and soft ones, and a feasible solution requires all hard clauses to be satisfied~\cite{li2021maxsat}. Among various complete and incomplete solvers developed for solving MaxSAT, stochastic local search (SLS)~\cite{CaiLLS16,LeiC18}, which commonly follows an iterative framework, is a significant category of incomplete solvers.
While new methods have been continuously proposed, the corresponding comparison results are presented in numerous documents. The famous MaxSAT Evaluations~\cite{bergmaxsat} provide a platform for practitioners to understand the performance of state-of-the-art solvers,
and the existing comparisons among MaxSAT solvers mainly concern the quality, i.e., fitness, of best solutions obtained within a given time budget, namely, cutoff time~\cite{luo2014ccls,CaiLTS14,CaiLuoZhang17,luo2017ccehc,AlKasemM21,ChuCL2023,LiuEtAl25}, which is a reasonable option when we require a concrete judgment in competitions. 
However, the assessments concerning best-found solutions may produce bias in the cutoff time and hinder us from understanding the behavior of solvers during the iterative optimization process.
Though some work~\cite{Zheng0Z0LM22} investigated the convergence process when analyzing SLS solvers' behavior, the current mechanism usually plots individual convergence lines for particular instances. Hickey and Beacchus's work~\cite{HickeyB22} addressed anytime performances of MaxSAT solvers. Still, their work focused on proposing a solution that would obtain good solutions on multiple cutoff times without addressing the issue of measuring anytime performance. Meanwhile, recent work~\cite{HansenABT22} has discussed that anytime performance assessment is vital for benchmarking algorithms.

Two perspectives, i.e., fixed-budget and fixed-target, are commonly considered when measuring iterative algorithms' performance.
The fixed-budget approach is usually applied to problems in which prior knowledge about the performance, e.g., the fitness scale, is unavailable, and it can assess the order of the obtained solutions. 
The fixed-target approach is commonly applied in well-understood practical scenarios or benchmarking environments, and it uses quantitative indicators to explain how fast an algorithm obtains a solution meeting specific requirements.
However, we usually need to consider both perspectives when developing an algorithm for practical problems.
Therefore, addressing the concerns of both approaches, an anytime performance metric, which examines algorithms' performance based on empirical cumulative distribution functions (ECDF) of a set of cutoff times, has recently presented its superiority in black-box optimization benchmarking scenarios~\cite{DoerrYHWSB20,HansenABTT16,HansenABT22}. The ECDF value at a given time is computed based on the fraction of solved problem instances or obtained solutions reaching the required target, e.g., fitness. In this way, we can obtain a ratio scaled value independent of the scale of solution fitness, and results regarding multiple cutoff times can be reasonably aggregated.

This paper introduces the first anytime performance assessments of MaxSAT SLS solvers. Our anytime performance assessments reveal intriguing behaviors that were invisible for fixed-budget assessments. The observations can provide valuable insights for improving the design and parameter settings of future SLS solvers. The hybrid solvers, which are the current state-of-the-art MaxSAT solvers, operate complete solvers and SLS solvers independently and alternatively~\cite{CaiL20}. Recent work on improving hybrid solvers mainly works on enhancing the performance of the SLS part~\cite{MSESOLVER}. Therefore,
this work focusing on analyzing SLS will certainly contribute to future enhancement of the state-of-the-art MaxSAT solvers. 
In addition, the assessment techniques employed in this work can be transparent to both complete and hybrid solvers.
In practice, our results \emph{reveal (dis)advantages of the solvers across different problem instances} and \emph{demonstrate algorithms' convergence progress aggregated across multiple instances}. Apart from providing a comprehensive assessment that can inspire experts to improve the design of algorithms further, the quantitative assessment also creates an alternative way for algorithmic tools, e.g., hyperparameter optimizers, to recognize algorithms' performance.

Since contemporary algorithms, including SLS, are mostly parametric, hyperparameter optimization (HPO) is vital for robust and competitive performance. For example, the MaxSAT Evaluation competition\footnote{\url{https://maxsat-evaluations.github.io/2023/}} submissions are usually embedded with fitting parameter settings. Meanwhile, automatic HPO tools have been commonly applied to achieve those fine-tuned settings. The HPO scenario for MaxSAT solvers usually addresses the algorithm configuration (AC) problem, which aims to find a configuration minimizing a cost function $e$. A configuration will be executed to evaluate the cost function $e$ until a cutoff time $\kappa$. Based on theoretical analysis of evolutionary algorithms, recent work~\cite{HallOS22} has studied the impact of $\kappa$ on the performance of HPO, considering both scenarios using fixed-target and fixed-budget approaches as $e$. Tuning algorithms' anytime performance has been addressed in the recent work~\cite{YeDWB22}, which compared the HPO scenarios of tuning fixed-target and anytime performances. While the mentioned studies were tested on classic benchmark problems such as \textsc{OneMax}~\cite{DoerrYHWSB20}, L{\'o}pez-Ib{\'a}{\~n}ez and St{\"u}tzle applied the hypervolume considering two objectives, i.e., the best-found solution quality and the corresponding cutoff time, to take into account the anytime behavior of algorithms~\cite{Lopez-IbanezS14}, and the method was tested on a MAX-MIN Ant system for Traveling Salesperson Problems and the famous mixed-integer solver SCIP~\cite{Achterberg09}. In this work, we investigate tuning the anytime performance of a MaxSAT SLS solver, and the experimental results suggest that \emph{anytime performance is a better mechanism than the fixed-budget one for the cost function of HPO}.

The main contributions of this work are as follows:
\begin{itemize}
    \item We provide an anytime performance assessment of four state-of-the-art MaxSAT SLS solvers. We aggregate their ECDFs across multiple problem instances and compare the solvers' performance concerning various cutoff times. The assessment illustrates that ECDF, as a universal technique with a ratio scale, can distinguish the solvers that are considered identical when comparing the fixed-budget performance. Moreover, we observe that the solvers' performance varies along the optimization process while investigating their performance in particular instances, addressing the necessity of anytime performance assessment.
    \item We suggest tuning anytime performance can obtain better parameter settings compared to the commonly applied fixed-budget performance tuning in the MaxSAT community.
    By tuning the ECDF of SLS solvers, SMAC~\cite{LindauerEFBDBRS22} can obtain better configurations in terms of anytime performance (i.e., ECDFs) and fixed-budget performance (i.e., scores based on the best-found solution quality). Note that the mechanism of computing incremental aggregated ECDFs for HPO used in this paper can be applied to other scenarios without interfacing with the HPO framework.
\end{itemize}
\section{Preliminaries}
\label{sec:pre}
\paragraph{Problem Definition}
Given a set of $n$ Boolean variables $V = \{x_1, x_2, \ldots, x_n\}$, a \emph{literal} $l$ is either a variable $x$ or its negation $\neg x$, and a \emph{clause} is a disjunction of literals, i.e., $c = l_1 \lor l_2 \ldots \lor l_k$, where $k$ is the length of $c$. Given an assignment $\alpha$ mapping Boolean values to each variable in $V$, a clause $c$ is satisfied if at least one literal in $c$ is \emph{True}; otherwise, it is falsified. Given a conjunction normal form (CNF) $F = c_1 \land c_2 \land \ldots \land c_m$, where $m$ is the number of clauses, the MaxSAT problem is to find an assignment that maximizes the number of satisfied clauses. Partial MaxSAT (PMS) is a variant of MaxSAT that divides clauses into \emph{hard} ones and \emph{soft} ones, and a feasible assignment $\alpha$ requires all \emph{hard} clauses to be satisfied. For the weighted PMS (WPMS) problem, each clause is associated with a weight $w(c) > 0$, and the problem is to find a feasible $\alpha$ that maximizes the total weights of satisfied clauses (namely, minimizes the total weights of falsified clauses). In this paper, we denote (W)PMS as a constrained optimization problem minimizing the $cost(\alpha)$ that is the total weight of falsified clauses. Nowadays, MaxSAT solvers are commonly developed and tested for (W)PMS, and we refer to MaxSAT as (W)PMS.

\paragraph{Algorithms}
SLS is a popular category of incomplete algorithms for MaxSAT. The common procedure of SLS samples an initial solution, i.e., assignment $\alpha$, and then follows an optimization loop, creating new assignments by flipping one or multiple variables $x$ in $\alpha$ iteratively until reaching the termination condition (e.g., using out the cutoff time). Variants of SLS solvers usually work on the strategies of picking one or multiple variable(s) $x$ to be flipped.

The classic SATLike solver~\cite{LeiC18} proposed a clause weighting scheme, which introduces a bias towards certain variables $x$ and determines which one(s) to be flipped. The newest extension of SATLike, SATLike3.0~\cite{CaiL20}, is one of the state-of-the-art SLS solvers for MaxSAT and has been included in much work for comparisons~\cite{AlKasemM21}. NuWLS~\cite{ChuCL2023} proposed a new clause weighting scheme handling hard and soft clauses separately and worked on the weight initialization based on the framework of SATLike3.0. Differently from other methods that flip one variable at each iteration, MaxFPS~\cite{Zheng0Z23} applies a farsighted probabilistic sampling method flipping a pair of variables. Moreover, instead of using the clause weighting technique, BandMax~\cite{Zheng0Z0LM22} introduced the method of multi-armed bandit to select clauses to be satisfied, correspondingly deciding which variable to be flipped.

In this paper, we select SATLike3.0, BandMax, NuWLS, and MaxFPS for our first assessment of anytime performance of MaxSAT SLS solvers.  Due to the space issue, we denote SATLike as SATLike3.0 in the following text.

\section{Performance Assessment}
\label{sec:assess}
We test the four state-of-the-art SLS solvers on the 2022 and 2023 anytime tracks of MaxSAT Evaluations, including weighted (MSE23-w and MSE22-w) and unweighted MaxSAT (MSE23-uw and MSE22-uw). All the tested solvers are implemented in C++ and complied with g++ 9.4.0. All experiments are conducted in Ubuntu 20.04.4 with the AMD EPYC 7763 CPU and 1TB RAM. The cutoff time is set as $300s$ for each run following the suggestion of MaxSAT Evaluations, and we conduct a solver of ten independent runs for each instance. 
Additional detailed results of the cutoff time $60s$, which is another suggestion of MaxSAT Evaluations, are available in our public repository.

We introduce in the following the solvers' fixed-budget performance and anytime performance. The former is commonly applied in the MaxSAT community, and the latter is studied for MaxSAT for the first time in this work. The fixed-target performance is not addressed because MaxSAT is an NP-hard problem, and its optima are unknown.
\subsection{Fixed-budget Performance}
\label{sec:fixedbudget}
\paragraph{Fixed-budget Performance Metric} Existing work on SLS solvers for MaxSAT usually reports their experimental results based on (1) the number of winning instances (``\#win'') and (2) the score based on the competing solution~\cite{bergmaxsat,AlKasemM21}. The number of winning instances presents the scope in the tested instances where a solver obtains the best solution among all competing solvers, and the score denotes the ratio between the solution quality obtained by a solver and the best-competing solver:
\begin{equation}
    score(i) = \frac{\text{cost of the best solution for } i + 1}{\text{cost of the found solution for } i + 1 } \in [0,1]
\label{eq:score}
\end{equation}
$score(i) = 0$ indicates that a solver can not find a feasible solution for the instance $i$. 
These two metrics are both calculated based on the best-found solution within a cutoff time. Consequently, such assessment outcomes may deviate along different cutoff times. Note that we are not critical of using these measures, as they can provide a concrete comparison for algorithm performance orders, as shown in amounts of existing work and MaxSAT Evaluation competitions.

\paragraph{Experimental Results} We present in Table~\ref{tab:winins} the number of instances in which a solver obtains the best solution (that is achieved by all runs of the competing solvers) in one of the tested ten runs, followed by the number (in brackets) of instances in which a solver obtains the best average results of ten runs among the competing solvers. Note that we omit numbers in brackets if they are identical to the former ones. Results are categorized for the four tested instance sets. According to Table~\ref{tab:winins}, NuWLS obtains significant advantages against the other solvers, and BandMax and MaxFPS show similar performance across the four tested benchmark tracks. BandMax and MaxFPS win on more instances than SATLike when comparing the best of ten runs, but such advantages diminish when comparing the average of ten runs (as shown in the brackets). This observation indicates the necessity of performing multiple trials when comparing SLS solvers.
\begingroup
\setlength{\tabcolsep}{4.8pt}
\begin{table}[t]
  \centering
    \begin{tabular}{lcccc}
\hline
 & BandMax & MaxFPS & NuWLS & SATLike \\
\hline
 MSE23-w & 42(30)  & 26(20) & 81& 26(18) \\
 MSE23-uw & 73(42) & 71(34) & 110(99) & 51(30)\\
 MSE22-w & 49(40) & 49(41)& 103 & 35(27)\\
 MSE22-uw & 89(65) & 86(64)& 106(91)& 55(39)\\
\hline
\end{tabular}
\caption{Number of instances in which a solver obtains the best performance among the competing solvers}
\label{tab:winins}
\end{table}
\endgroup

\begin{figure*}[t]
    \centering
    \includegraphics[trim=0 10 0 5, clip, width=\linewidth]{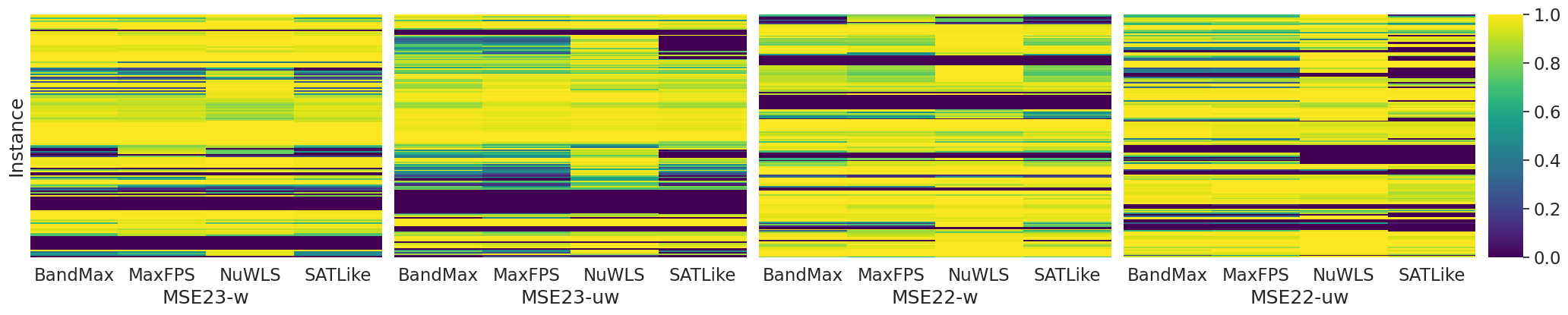}
    \caption{
    Heatmap illustrating the scores for individual instances. Each row represents an instance, while each column represents a different solver. The color depicts the score achieved by the solvers, with lighter shades indicating better performance. The benchmark tracks ``MSE23-w'', ``MSE23-uw'', ``MSE22-w'' and ``MSE22-uw'' consist of 160, 179, 197, and 179 instances (as shown by four boxes) respectively. The names of the instances have been omitted from the y-axis due to the limited space. \revise{Detailed results regarding groups of instances are available in Appendix B.}
    }
    \label{fig:heatscore}
\end{figure*}

Table~\ref{tab:scores} presents the scores of the tested solvers, of which values are averaged across the instances for each benchmark track. Since we conduct ten independent runs for each instance, the score of a solver for an instance $i$ is the average of ten trails. $score(i) = 1$ indicates that a solver obtains the best solution for the instance $i$ among the competing solvers in all runs, and $score(i) = 0$ indicates that no feasible solution is obtained. We do not take the scores of the instances in which no solver obtains a feasible solution into account.  We observe that the orders of the solvers' scores are identical to the ones in Table~\ref{tab:winins}. In addition, according to Figure~\ref{fig:heatscore}, there exist several instances in which no solver can obtain a feasible solution, and the solvers obtain the same or similar results in many instances, as shown in light yellow color, e.g., $score > 0.9$. 
Also, Figure~\ref{fig:heatscore} shows NuWLS obtains better scores in more instances compared to the other solvers.
 
\begingroup
\setlength{\tabcolsep}{4.8pt}
\begin{table}[t]
\begin{tabular}{lcccc}
\hline
 & BandMax & MaxFPS & NuWLS & SATLike \\
\hline
MSE23-w & 0.849 & 0.867 & 0.904 & 0.827 \\
MSE23-uw & 0.800 & 0.782 & 0.902 & 0.705 \\
MSE22-w & 0.869 & 0.889 & 0.942 & 0.864 \\
MSE22-uw & 0.854 & 0.852 & 0.901 & 0.731 \\
\hline
\end{tabular}
\caption{Aggregated scores of the solves}
\label{tab:scores}
\end{table}
\endgroup

\subsection{Anytime Performance}
The assessment of fixed-budget performance provides initial insights into the performance of the tested solvers. However, these results only focus on the cutoff time of $300s$. While we can conduct similar analyses for multiple cutoff times,
for example, the MaxSAT Evaluations platform includes cutoff times of $60s$ and $300s$, 
it remains challenging to aggregate and interpret the fixed-budget results across multiple cutoff time scenarios to understand the convergence of solvers. Therefore, we ask for an indicator that can assess anytime performance across various time budgets. Furthermore, this indicator shall be quantitative, allowing us to measure performance differences on a scale rather than by order. Particularly for instance-based scenarios such as MaxSAT, this indicator needs to be \emph{universal} and independent of the scale of solution quality (e.g., cost of assignments) for each instance, enabling fair performance aggregation across instances.

\begin{figure}[htb]
    \centering
    \includegraphics[trim=10 12 0 12, clip,width=\linewidth]{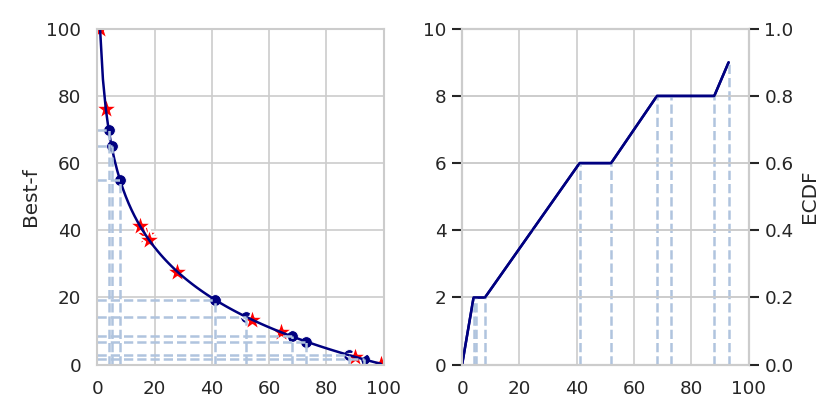}
    \caption{An example of ECDF calculation. \textbf{Left:} The blue dots represent pairs $(t,\phi)$, where a better solution with fitness $\phi$ ($y$-axis) is obtained at time $t$ ($x$-axis). \textbf{Right:} the ECDF values based on the given set of solution fitness as presented in red stars on the left.}
    \label{fig:calecdf}
\end{figure}

\paragraph{Anytime Performance Metric.} 
We assess the anytime performance of MaxSAT solvers using the ECDF values of a set of cutoff times. ECDF indicates the fraction of the obtained solutions satisfying a specific quality.
We denote $\phi$ as the solution quality, e.g., the cost of an assignment.
Given a set of solutions with $\Phi_i = \{\phi_{i1}, \phi_{i2}, \ldots\}$ for a problem instance $i$ and a solver $A$ that obtains the best-found solution with $\phi_{A_i}$ within the cutoff $t$, $A$'s ECDF value at $t$ is:
\begin{equation}
    \text{ECDF}(A, i, t) = \frac{\mid \{\phi \in \Phi_i \mid \phi \ge \phi_{A_i}\} \mid}{\mid \Phi_i \mid}
\end{equation}
We work on minimization in this paper,  and $ \{\phi \in \Phi_i \mid \phi \ge \phi_{A_i}\}$ denotes the subset of solutions that are not better than the best-found one of $A$ for instance $i$. 
For example, an algorithm $A$ obtains a set of pairs of optimization time and the corresponding best-found solution $O= \{(t_1, \phi_{o1}), (t_2,\phi_{o2}),\ldots\}$, which are plotted by blue dots in Figure~\ref{fig:calecdf}. Given a set $\Phi$ of solutions plotted by red stars, we present on the right subfigure the number of instances in $\Phi$ that are not better than the corresponding $\phi_o$ for each $t$, and the values normalized by the size of $\Phi$ are ECDFs.

Because ECDF is with a ratio scale, we can evaluate the ECDFs of a solver $A$ for a set of optimization times $t$ to measure $A$'s \emph{anytime} performance. ECDFs can also be combined across multiple problem instances $i$ by considering specific $\Phi$ for each instance.

To calculate ECDFs, we form a set of solution fitness values $\Phi_i$ for each instance $i$. This solution fitness set comprises the costs of solutions that the solvers have visited during our experiments. We compute the ECDFs at $100$ specific optimization times, chosen within the range of $[0,300s]$ following a logarithmic scale. This set of optimization times introduces a preference for evaluating the performance in terms of attaining high-quality solutions in less time.

Note that measuring ECDFs has been adopted in the black-box optimization community, where researchers quantify ``time'' in terms of function evaluations, i.e., the number of solutions that have been evaluated~\cite{HansenABTT16,WangVYDB22}. Function evaluation ensures reproducibility and erases disturbances caused by hardware environment, programming design, and other factors. However, MaxSAT solvers are usually applied in practical scenarios concerning cpu time, and cpu time has been commonly used when measuring MaxSAT solvers' performance. We have conducted analyses by using cpu time and function evaluations as the time metrics, respectively. The analyses using these two time metrics show identical conclusions. Therefore, we present the results using cpu time in this paper. The data using function evaluations that are supported in a favored black-box benchmarking platform~\cite{WangVYDB22,de2023iohexperimenter} is available in our repository.

\begin{figure}[t]
\centering
\subcaptionbox{\textbf{Left:} WPMS (instances of MSE22-w and MSE23-w) and \textbf{Right:} PMS (instances of MSE22-uw and MSE23-uw)\label{fig:any-ecdf1}}{\includegraphics[trim=10 10 0 10, clip, width=\linewidth]{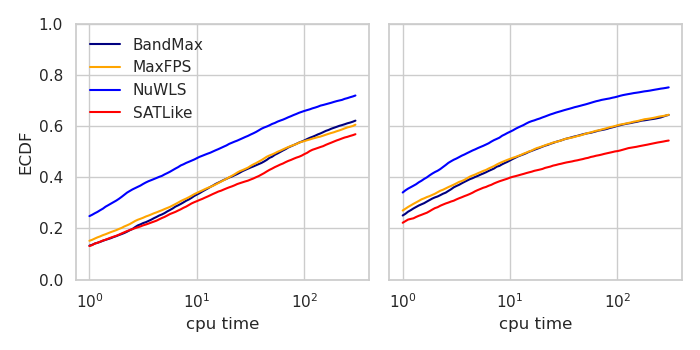}}

\subcaptionbox{ \textbf{Left:} ``decision-tree'' instances ($15$ in total), and \textbf{Right:} ``ParametricRBACMaintenance'' instances ($13$ in total).\label{fig:any-ecdf2}}{\includegraphics[trim=10 10 0 10, clip, width=\linewidth]{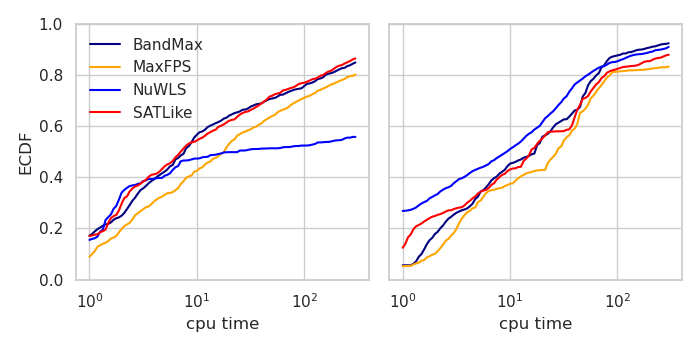}}
\caption{Aggregated ECDFs of each solver for different sets of problem instances. $x$-axis indicates cpu time, and $y$-axis indicates the corresponding aggregated ECDF values.}
\label{fig:any-ecdf}
\end{figure}

\paragraph{Observing Anytime Performance.} Figure~\ref{fig:any-ecdf1} shows that NuWLS outperforms the other solvers for both weighted and unweighted MaxSAT. BandMax and MaxFPS show similar performance, but by examining the ``anytime'' analysis, we can see that MaxFPS slightly outperforms BandMax within a time limit of $10s$, and the order of their performances alter as the optimization time increases for the weighted MaxSAT. Furthermore, detailed analysis of anytime performance for specific instances can reveal exciting findings. Although NuWLS demonstrates advantages over other solvers considering ECDFs aggregated across the entire set of tested problem instances, it does not exhibit superiority on ``decision-tree'' instances after being trapped in local optima around $10s$, as shown in Figure~\ref{fig:any-ecdf2}. On the other hand, while BandMax obtains similar performance to SATLike for ``ParametricRBACMaintenance'' instances when cpu time is less than $100s$, it outperforms the other solvers afterward. Compared to the fixed-budget performance assessment, these anytime performance observations offer more insights into the behavior of algorithms for specific problem instances and can be valuable for enhancing algorithm design.

\begin{figure*}[t]
    \centering
    \includegraphics[trim=0 10 0 5, clip, width=\linewidth]{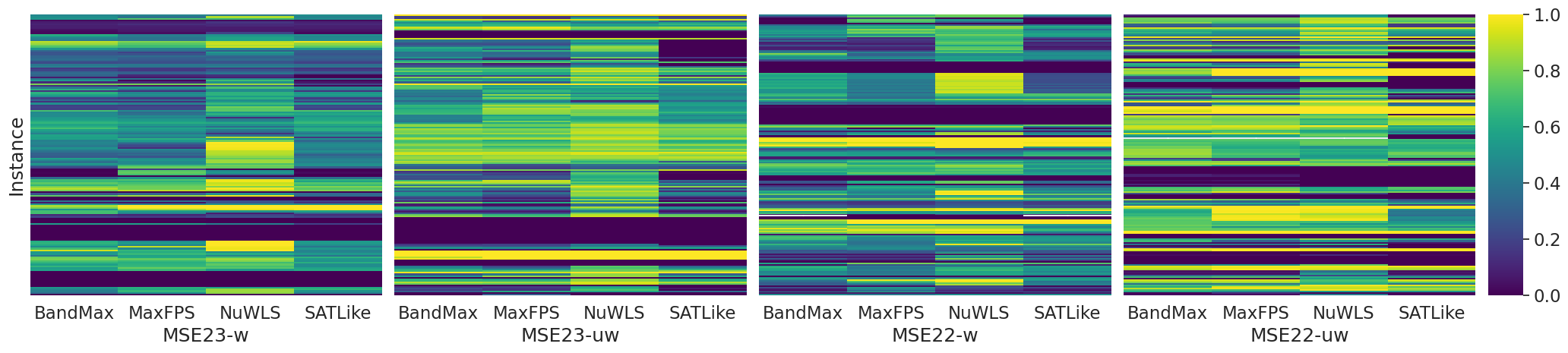}
    \caption{Heatmap illustrating aggregated ECDFs for individual instances. It follows the layout of Figure~\ref{fig:heatscore}, but the color depicts the ECDF values achieved by the solvers. \revise{Detailed results regarding groups of instances are available in Appendix B.}}
    \label{fig:ecdf}
\end{figure*}

\begingroup
\setlength{\tabcolsep}{4.8pt}
\begin{table}[t]
    \centering
    
    \begin{tabular}{lcccc}
\hline
 & BandMax & MaxFPS & NuWLS & SATLike \\
\hline
MSE23-w & 0.444 & 0.445 & 0.584 & 0.409 \\
MSE23-uw & 0.546 & 0.548 & 0.696 & 0.466 \\
MSE22-w & 0.458 & 0.470 & 0.624 & 0.420 \\
MSE22-uw & 0.589 & 0.602 & 0.696 & 0.494 \\
\hline
\end{tabular}
    \caption{Aggregated ECDFs for each benchmark track. The ECDF value for each run on an instance is normalized by the number of considered cutoff times, i.e., $100$. The presented values are average across the corresponding instances. }
    \label{tab:ecdf}
\end{table}
\endgroup
\paragraph{Quantitative Assessment with Variation}
As mentioned previously, ECDF has a ratio scale that allows aggregation of values across multiple optimization times to obtain quantitative assessments for each instance, which is the so-called (approximated) area under the ECDF curve (AUC) \cite{YeDWB22}.
Table~\ref{tab:ecdf} presents the aggregated ECDFs for the four benchmark tracks, showing that the order of the solver's assessment based on ECDF remains the same as the one based on scores, as presented in Table~\ref{tab:scores}. When analyzing the behavior in specific instances, Figure~\ref{fig:ecdf} shows higher variances in ECDFs for each instance compared to the ones in scores (see Figure~\ref{fig:heatscore}). Specifically, instead of having a large set of scores higher than $0.9$ in Figure~\ref{fig:heatscore}, ECDFs are in a wide variation of values as shown in Figure~\ref{fig:ecdf}. \revise{We also provide the average standard deviations of scores and ECDFs across all tested instances in Appendix A, which shows that the standard deviations obtained from ECDFs are much greater than the scores.}
Moreover, for the instances where solvers can obtain the same best-found solutions, ECDFs can distinguish solvers' performance by quantitatively estimating the convergence process. For example, Figure~\ref{fig:ecdf} can show that NuWLS converge faster for the instances plotted in yellow.

\section{Hyperparameter Optimization}
Contemporary algorithms, such as heuristics, evolutionary algorithms, machine learning models, etc., are usually parametric. Therefore, parameter settings are essential for the competitive performance of algorithms. Before the development of automatic tools, parameter settings were manually chosen based on experts' experience or repetitive testing using Grid Search. Nowadays, the tools such as SMAC~\cite{LindauerEFBDBRS22} and irace~\cite{lopez2016irace} have been commonly applied for automatic HPO. While practical scenarios may propose various requirements for HPO, we address in this paper the algorithm configuration (AC) problem, which meets the requirement of tuning MaxSAT solvers for a set(s) of problem instances. AC~\cite{LindauerEFBDBRS22} aims at searching a well-performance configuration, i.e., parameter setting, $\lambda \in \Lambda$ of an algorithm $A$ across a set of problem instances $\{i_1, \ldots\} \subset \mathcal{I}$:
\begin{equation}
\mathbf{\lambda}^* \in \underset{\lambda \in \Lambda}{\arg \min} \  e(\lambda)
\label{eq:ac}
\end{equation}
where the cost function $e$ is usually set as the best-found solutions' quality intuitively when tuning MaxSAT SLS solvers. However, in this section, we introduce AUC as an alternative candidate for $e$ and compare the results obtained by using these two different mechanisms as the cost function for tuning MaxSAT SLS solvers. For ease of reading, we denote AUC and aggregated ECDFs as the metrics used in HPO and experimental comparisons, respectively.

\paragraph{Settings of Hyperparameter Optimization.} We use the AC arcade of SMAC3~\footnote{https://github.com/automl/SMAC3} to compare the results of tuning fixed-budget performance and anytime performance.
\revise{\textbf{Best-f}, which calculates the fitness of the best-found solution obtained within a given cutoff $t_B$, is a straightforward option for the cost function of tuning fixed-budget performance. In addition, we apply $-score$ in~Eq.~\ref{eq:score}, which is denoted as \textbf{Norm-f}, as another cost function of tuning fixed-budget performance. To calculate scores, we use the best-found solutions obtained in the Performance Assessment section. Note that this function can be adopted for competitions but is not doable for the practical scenarios lacking baselines, i.e., the known best-found solutions.}
As for calculating the cost function of anytime performance \textbf{approximated AUC}, we form a set $T$ of $50$ optimization times chosen within a range of $[0.1s,t_B]$ following a logarithmic scale. 
Since the time budget $t_B$ for executing each run of configurations is usually relatively small for HPO scenarios, e.g., $t_B = 100s < 300s$ in this work, following a bias toward the configurations that have the potential to yield better solutions when provided with an additional time budget, we use $t_B - T$ as the set of optimization times to calculate AUCs. Regarding the target (i.e., solution quality) set $\Phi_i$ for each instance $i$, we start with a set $F_i$  of five fitness values chosen from $[f_\text{init\_min},f_\text{init\_max}]$ based on the linear scale, where $f_\text{init\_min}$ and $f_\text{init\_max}$ are the fitness of the best and the worst solutions visited by the initial configuration. When a configuration obtains a better best-found solution for an instance during the tuning process, the corresponding best-found fitness value will be added to $F_i$. 
In addition, the AUC of a configuration for a given instance is calculated by aggregating the values across $50$ optimization times.
In practice, we use a modified $\text{AUC}' =  - \text{AUC } \cdot \mid \Phi_i \mid$ measuring $e$ of configurations for each instance $i$. $\text{AUC}'$ decreases accordingly when a better configuration achieves a better best-found solution, thereby increasing the size of $\Phi$. In this way, all of the tuning scenarios are framed as minimization tasks for HPO. Though computing AUC is more complex than computing Best-f and Norm-f, in practice, this computing takes linear time, and its time consumption is negligible compared to the time budget allocated for each run of a configuration.

\revise{We have tuned the four solvers for the MSE23-w and MSE23-uw benchmark tracks by using SMAC with three different cost functions.}
We conduct five runs of each SMAC setting with the cpu time budget of $60,000s$, following the suggestion in the work of NuWLS~\cite{ChuCL2023}. Twenty percent of the instances are randomly selected for training.  
The obtained configurations of solvers are compared based on scores and AUCs across all the instances. Note that the computation of scores and ACUs follows the same settings as the Performance Assessment section.

\begin{table*}[htb]
    \centering
{    
    \begin{tabular}{r|ccc|ccc|ccc|ccc}
\hline
 & \multicolumn{3}{c}{BandMax} & \multicolumn{3}{c}{MaxFPS} & \multicolumn{3}{c}{NuWLS} & \multicolumn{3}{c}{SATLike} \\
\hline
 & Best-f & Norm-f &  ECDF & Best-f &  Norm-f & ECDF & Best-f  &  Norm-f & ECDF & Best-f &  Norm-f & ECDF \\
\hline
\multicolumn{13}{c}{Results on MSE23-w}\\
\hline
Score&	0.857	&	0.850	&	\textbf{0.858}	&	0.844	&	\textbf{0.850}	&	0.848	&	0.795	&	0.809	&	\textbf{0.818}	&	0.664	&	0.564	&	\textbf{0.758} \\
ECDF	&	\textbf{0.423} 	&	0.405 	&	\textbf{0.423} 	&	0.424 	&	0.423 	&	\textbf{0.430} 	&	0.431 	&	0.450 	&	\textbf{0.479} 	&	0.288 	&	0.225 	&	\textbf{0.353}  \\
\hline
\multicolumn{13}{c}{Results on MSE23-uw}\\
\hline
Score &	\textbf{0.815}	&	0.799	&	0.814	&	\textbf{0.776}	&	0.773	&	0.769	&	0.824	&	0.827	&	\textbf{0.846}	&	0.617	&	0.717	&	\textbf{0.790}	\\
ECDF&	\textbf{0.549} 	&	0.532 	&	0.548 	&	\textbf{0.505} 	&	0.503 	&	0.496 	&	0.488 	&	0.491 	&	\textbf{0.528} 	&	0.373 	&	0.423 	&	\textbf{0.477}  \\
\hline
\end{tabular}

}
\caption{Average scores and ECDFs of five configurations obtained by each setting. Larger values indicate better performances. 
}
\label{tab:tuned}
\end{table*}

\begin{figure}[htb!]
    \centering
    \subcaptionbox{BandMax}{\includegraphics[trim=5 5 0 5, clip,width=\linewidth]{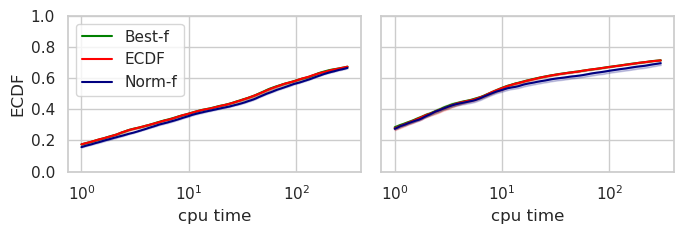}}
     \subcaptionbox{MaxFPS}{\includegraphics[trim=5 5 0 5, clip,width=\linewidth]{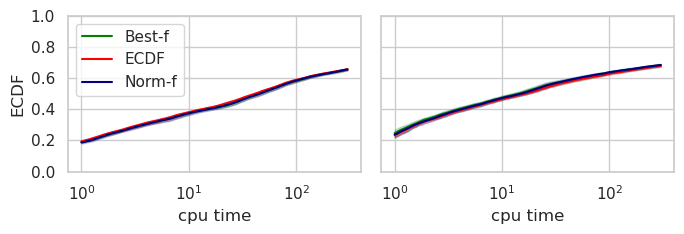}}
    \subcaptionbox{NuWLS}{\includegraphics[trim=5 5 0 5, clip,width=\linewidth]{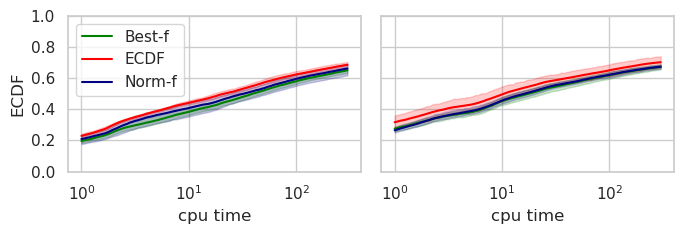}}
     \subcaptionbox{SATLike}{\includegraphics[trim=5 5 0 5, clip,width=\linewidth]{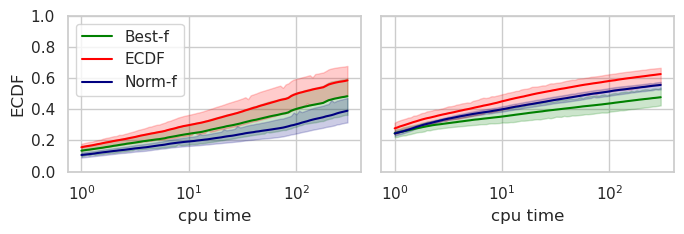}}

    \caption{Aggregated ECDFs of the configurations obtained by tuning for Best-f, Norm-f, and ECDF for \textbf{Left:} MSE23-w and \textbf{Right:} MSE23-uw. Results plotted in one line are from five configurations obtained by independent runs of SMAC.}
    \label{fig:tunedecdf}
\end{figure}

\paragraph{Anytime Performance can Lead to Better Configurations.}
Table~\ref{tab:tuned} presents the results of configurations obtained by 
each setting of SMAC, e.g., tuning solvers using different cost functions, and Figure~\ref{fig:tunedecdf} demonstrates the mean and $95\%$ confidence interval of aggregated ECDFs of five configurations along cpu time. More detailed results of each tuned configuration are provided in Appendix C, and all results are from $120$ $(4 \text{ solvers} \times 3 \text{ cost functions} \times 2 \text{ benchmark tracks} \times 5 \text{ SMAC runs})$ configurations.

\revise{We first compare the anytime performance of the obtained configurations for eight scenarios (4 solvers $\times$ 2 benchmark tracks).
According to Table~\ref{tab:tuned}, using AUC as the cost function can obtain the best anytime performance for six out of eight scenarios, except for BandMax and MaxFPS in MSE23-uw.
We can observe in Figure~\ref{fig:tunedecdf}a-b that the anytime performance of the tuned BandMax and MaxFPS configurations are close to each other, and the results of using AUC as the cost function generally obtain slight advantages along CPU time for BandMax. 
When comparing the fixed-budget performance, i.e., scores, using AUC as the cost function obtains the best results for five out of eight scenarios in Table~\ref{tab:tuned}. Note that Wilcoxon signed-rank tests indicate that the results of using AUC as the cost function significantly differ from the others. P-values are provided in Appendix C.}

\revise{Overall, our results suggest that tuning anytime performance can lead to better configurations in terms of both anytime performance and best-found solution quality. Specifically, when tuning NuWLS and SATLike, using AUC as the cost function obtains significant advantages ($4\%$ and $15\%$ by average) compared to the second-best results.
For the five out $16$ scenarios (in Table~\ref{tab:tuned}) that tuning AUC does not obtain the best results, the obtained results are only $0.6\%$ worse than the best ones by average. }

\paragraph{Reasons for the Better Results.} One reason for the findings is that AUCs can provide \emph{dense search space} for HPO.
For example, within a given time budget $t_B$, when two configurations $\lambda_1$ and $\lambda_2$ obtain the same best-found solution for an instance $i$ using different time $t \le t_B$, Best-f and Norm-f will deliver identical information to HPO tools, but AUC can distinguish these two configurations.
Another reason is that AUCs are \emph{normalized values} ranging between $0$ and $1$ compared to Best-f. Although Norm-f is also normalized, its values range in a smaller domain for each instance. In contrast, AUCs have a wider variation of values.
\revise{Note that we have also tested using the average of AUCs and Norm-f values across 20 instances as the cost function each time when SMAC evaluates a configuration. However, according to the results in Appendix D, normalization across multiple instances does not help configure MaxSAT solvers.}
In addition, when conducting HPO, the time budget $t_B$ is usually set as a relatively small value due to time limits. For example, in our tuning process of SMAC, we allocated a budget of $100s$ for each configuration. However, when validating the obtained configurations, the budget is $300s$. 
In the scenarios where configurations are allocated with limited time budgets, by considering multiple cutoff times, AUCs can \emph{robustly estimate} the potential ability of a configuration to achieve better solutions. On the other hand, using Best-f or Norm-f as the cost function may be misled by fortunate results obtained at a specific cutoff time.

\section{Conclusion}
We have introduced ECDF for anytime performance analysis of MaxSAT SLS solvers. Our assessments have been conducted for the four state-of-the-art solvers, namely BandMax, MaxFPS, NuWLS, and SATLike. We illustrate that ECDF can measure solvers' anytime performance regarding multiple cutoff times, provide quantitative assessments with a ratio scale that can be aggregated across multiple instances, and differentiate the solvers that are considered similar in terms of fixed-budget performance.
\revise{Our experiments show that the AUC can serve as a competitive metric for hyperparameter optimization. Compared to the traditional fixed-budget cost functions, the AUC can help achieve configurations that are, on average, around $10\%$ better than the second-best ones in 11 out of 16 tested scenarios. In the remaining scenarios, its results are, on average, only $0.6\%$ worse than the best ones.}
Although we have presented the application in the context of MaxSAT SLS, we expect that the applied techniques can be transparent to the hybrid methods integrating complete and incomplete solvers. 
Furthermore, we have observed that the choice of cost functions can affect the results of HPO. Therefore, we plan to study multi-objective HPO and explore the effect of cost functions. Also, we will study the algorithm portfolio configuration by considering the solvers' anytime performance. 


\newpage
\section{Acknowledgements}
This work is supported by the National Key R\&D Program of China (2023YFA1009500), the National Natural Science Foundation of China (62202025), the Beijing Natural Science Foundation (L241050), Young Elite Scientist Sponsorship Program by CAST (YESS20230566), CCF-Huawei Populus Grove Fund (CCF-HuaweiFM2024005), and Fundamental Research Fund Project of Beihang University.
\bibliography{aaai25}

\begin{thebibliography}{28}
\providecommand{\natexlab}[1]{#1}

\bibitem[{Achterberg(2009)}]{Achterberg09}
Achterberg, T. 2009.
\newblock {SCIP:} solving constraint integer programs.
\newblock \emph{Mathematical Programming Computation}, 1(1): 1--41.

\bibitem[{AlKasem and Menai(2021)}]{AlKasemM21}
AlKasem, H.~H.; and Menai, M. E.~B. 2021.
\newblock Stochastic local search for Partial Max-SAT: an experimental evaluation.
\newblock \emph{Artificial Intelligence Review}, 54(4): 2525--2566.

\bibitem[{Berg et~al.(2023{\natexlab{a}})Berg, J{\"a}rvisalo, Martins, and Niskanen}]{bergmaxsat}
Berg, J.; J{\"a}rvisalo, M.; Martins, R.; and Niskanen, A. 2023{\natexlab{a}}.
\newblock MaxSAT Evaluation 2023.
\newblock \emph{Department of Computer Science Series of Publications B}, B-2023-2.

\bibitem[{Berg et~al.(2023{\natexlab{b}})Berg, J\"arvisalo, Martins, and Niskanen}]{MSESOLVER}
Berg, J.; J\"arvisalo, M.; Martins, R.; and Niskanen, A. 2023{\natexlab{b}}.
\newblock {MaxSAT Evaluation 2023 : Solver and Benchmark Descriptions0}.
\newblock \url{http://hdl.handle.net/10138/564026}.

\bibitem[{Biere, Heule, and van Maaren(2009)}]{biere2009handbook}
Biere, A.; Heule, M.; and van Maaren, H. 2009.
\newblock \emph{Handbook of satisfiability}, volume 185.
\newblock IOS press.

\bibitem[{Cai and Lei(2020)}]{CaiL20}
Cai, S.; and Lei, Z. 2020.
\newblock Old techniques in new ways: Clause weighting, unit propagation and hybridization for maximum satisfiability.
\newblock \emph{Artificial Intelligence}, 287: 103354.

\bibitem[{Cai et~al.(2016)Cai, Luo, Lin, and Su}]{CaiLLS16}
Cai, S.; Luo, C.; Lin, J.; and Su, K. 2016.
\newblock New local search methods for partial MaxSAT.
\newblock \emph{Artificial Intelligence}, 240: 1--18.

\bibitem[{Cai et~al.(2014)Cai, Luo, Thornton, and Su}]{CaiLTS14}
Cai, S.; Luo, C.; Thornton, J.; and Su, K. 2014.
\newblock Tailoring Local Search for Partial MaxSAT.
\newblock In Brodley, C.~E.; and Stone, P., eds., \emph{Proceedings of the Twenty-Eighth {AAAI} Conference on Artificial Intelligence}, 2623--2629.

\bibitem[{Cai, Luo, and Zhang(2017)}]{CaiLuoZhang17}
Cai, S.; Luo, C.; and Zhang, H. 2017.
\newblock From Decimation to Local Search and Back: {A} New Approach to MaxSAT.
\newblock In Sierra, C., ed., \emph{Proceedings of the Twenty-Sixth International Joint Conference on Artificial Intelligence}, 571--577.

\bibitem[{Chu, Cai, and Luo(2023)}]{ChuCL2023}
Chu, Y.; Cai, S.; and Luo, C. 2023.
\newblock NuWLS: Improving Local Search for (Weighted) Partial MaxSAT by New Weighting Techniques.
\newblock In Williams, B.; Chen, Y.; and Neville, J., eds., \emph{Proceedings of the Thirty-Seventh {AAAI} Conference on Artificial Intelligence}, 3915--3923.

\bibitem[{de~Nobel et~al.(2023)de~Nobel, Ye, Vermetten, Wang, Doerr, and B{\"a}ck}]{de2023iohexperimenter}
de~Nobel, J.; Ye, F.; Vermetten, D.; Wang, H.; Doerr, C.; and B{\"a}ck, T. 2023.
\newblock {IOHexperimenter}: Benchmarking platform for iterative optimization heuristics.
\newblock \emph{Evolutionary Computation}, 1--6.

\bibitem[{Doerr et~al.(2020)Doerr, Ye, Horesh, Wang, Shir, and B{\"{a}}ck}]{DoerrYHWSB20}
Doerr, C.; Ye, F.; Horesh, N.; Wang, H.; Shir, O.~M.; and B{\"{a}}ck, T. 2020.
\newblock Benchmarking discrete optimization heuristics with IOHprofiler.
\newblock \emph{Applied Soft Computing}, 88: 106027.

\bibitem[{Hall, Oliveto, and Sudholt(2022)}]{HallOS22}
Hall, G.~T.; Oliveto, P.~S.; and Sudholt, D. 2022.
\newblock On the impact of the performance metric on efficient algorithm configuration.
\newblock \emph{Artificial Intelligence}, 303: 103629.

\bibitem[{Hansen et~al.(2016)Hansen, Auger, Brockhoff, Tusar, and Tusar}]{HansenABTT16}
Hansen, N.; Auger, A.; Brockhoff, D.; Tusar, D.; and Tusar, T. 2016.
\newblock {COCO:} Performance Assessment.
\newblock \emph{CoRR}, abs/1605.03560.

\bibitem[{Hansen et~al.(2022)Hansen, Auger, Brockhoff, and Tusar}]{HansenABT22}
Hansen, N.; Auger, A.; Brockhoff, D.; and Tusar, T. 2022.
\newblock Anytime Performance Assessment in Blackbox Optimization Benchmarking.
\newblock \emph{{IEEE} Transactions on Evolutionary Computation}, 26(6): 1293--1305.

\bibitem[{Hickey and Bacchus(2022)}]{HickeyB22}
Hickey, R.; and Bacchus, F. 2022.
\newblock Large Neighbourhood Search for Anytime MaxSAT Solving.
\newblock In Raedt, L.~D., ed., \emph{Proceedings of the Thirty-First International Joint Conference on Artificial Intelligence}, 1818--1824.

\bibitem[{Lei and Cai(2018)}]{LeiC18}
Lei, Z.; and Cai, S. 2018.
\newblock Solving (Weighted) Partial MaxSAT by Dynamic Local Search for {SAT}.
\newblock In Lang, J., ed., \emph{Proceedings of the Twenty-Seventh International Joint Conference on Artificial Intelligence}, 1346--1352.

\bibitem[{Li and Manya(2021)}]{li2021maxsat}
Li, C.~M.; and Manya, F. 2021.
\newblock MaxSAT, hard and soft constraints.
\newblock In \emph{Handbook of satisfiability}, 903--927. IOS Press.

\bibitem[{Lindauer et~al.(2022)Lindauer, Eggensperger, Feurer, Biedenkapp, Deng, Benjamins, Ruhkopf, Sass, and Hutter}]{LindauerEFBDBRS22}
Lindauer, M.; Eggensperger, K.; Feurer, M.; Biedenkapp, A.; Deng, D.; Benjamins, C.; Ruhkopf, T.; Sass, R.; and Hutter, F. 2022.
\newblock {SMAC3:} {A} Versatile Bayesian Optimization Package for Hyperparameter Optimization.
\newblock \emph{Journal of Machine Learning Research}, 23: 54:1--54:9.

\bibitem[{Liu et~al.(2025)Liu, Liu, Luo, Cai, Lei, Zhang, Chu, and Zhang}]{LiuEtAl25}
Liu, C.; Liu, G.; Luo, C.; Cai, S.; Lei, Z.; Zhang, W.; Chu, Y.; and Zhang, G. 2025.
\newblock Optimizing local search-based partial {MaxSAT} solving via initial assignment prediction.
\newblock \emph{Science China Information Sciences}, 68(2): 122101:1--122101:15.

\bibitem[{L{\'o}pez-Ib{\'a}{\~n}ez et~al.(2016)L{\'o}pez-Ib{\'a}{\~n}ez, Dubois-Lacoste, C{\'a}ceres, Birattari, and St{\"u}tzle}]{lopez2016irace}
L{\'o}pez-Ib{\'a}{\~n}ez, M.; Dubois-Lacoste, J.; C{\'a}ceres, L.~P.; Birattari, M.; and St{\"u}tzle, T. 2016.
\newblock The irace package: Iterated racing for automatic algorithm configuration.
\newblock \emph{Operations Research Perspectives}, 3: 43--58.

\bibitem[{L{\'{o}}pez{-}Ib{\'{a}}{\~{n}}ez and St{\"{u}}tzle(2014)}]{Lopez-IbanezS14}
L{\'{o}}pez{-}Ib{\'{a}}{\~{n}}ez, M.; and St{\"{u}}tzle, T. 2014.
\newblock Automatically improving the anytime behaviour of optimisation algorithms.
\newblock \emph{European Journal Of Operational Research}, 235(3): 569--582.

\bibitem[{Luo et~al.(2017)Luo, Cai, Su, and Huang}]{luo2017ccehc}
Luo, C.; Cai, S.; Su, K.; and Huang, W. 2017.
\newblock CCEHC: An efficient local search algorithm for weighted partial maximum satisfiability.
\newblock \emph{Artificial Intelligence}, 243: 26--44.

\bibitem[{Luo et~al.(2014)Luo, Cai, Wu, Jie, and Su}]{luo2014ccls}
Luo, C.; Cai, S.; Wu, W.; Jie, Z.; and Su, K. 2014.
\newblock CCLS: an efficient local search algorithm for weighted maximum satisfiability.
\newblock \emph{IEEE Transactions on Computers}, 64(7): 1830--1843.

\bibitem[{Wang et~al.(2022)Wang, Vermetten, Ye, Doerr, and B{\"{a}}ck}]{WangVYDB22}
Wang, H.; Vermetten, D.; Ye, F.; Doerr, C.; and B{\"{a}}ck, T. 2022.
\newblock {IOHanalyzer}: Detailed Performance Analyses for Iterative Optimization Heuristics.
\newblock \emph{{ACM} Transactions on Evolutionary Learning and Optimization}, 2(1): 3:1--3:29.

\bibitem[{Ye et~al.(2022)Ye, Doerr, Wang, and B{\"{a}}ck}]{YeDWB22}
Ye, F.; Doerr, C.; Wang, H.; and B{\"{a}}ck, T. 2022.
\newblock Automated Configuration of Genetic Algorithms by Tuning for Anytime Performance.
\newblock \emph{{IEEE} Transactions on Evolutionary Computation}, 26(6): 1526--1538.

\bibitem[{Zheng, He, and Zhou(2023)}]{Zheng0Z23}
Zheng, J.; He, K.; and Zhou, J. 2023.
\newblock Farsighted Probabilistic Sampling: {A} General Strategy for Boosting Local Search MaxSAT Solvers.
\newblock In Williams, B.; Chen, Y.; and Neville, J., eds., \emph{Proceedings of the Thirty-Seventh {AAAI} Conference on Artificial Intelligence}, 4132--4139.

\bibitem[{Zheng et~al.(2022)Zheng, He, Zhou, Jin, Li, and Many{\`{a}}}]{Zheng0Z0LM22}
Zheng, J.; He, K.; Zhou, J.; Jin, Y.; Li, C.; and Many{\`{a}}, F. 2022.
\newblock BandMaxSAT: {A} Local Search MaxSAT Solver with Multi-armed Bandit.
\newblock In Raedt, L.~D., ed., \emph{Proceedings of the Thirty-First International Joint Conference on Artificial Intelligence}, 1901--1907.

\end{thebibliography}

\newpage
\include{reproducibility}
\end{document}


\maketitle
\appendix
This appendix attaches additional results of the AAAI'25 publication \emph{Better Understandings and Configurations in MaxSAT Stochastic Local Search Solvers via Anytime Performance Analysis}.

\section{A. Standard Deviations of the Solvers' Performance}
Referring to the results in Figure~1 and Figure~4 in the submission, we provide here the standard deviations of the scores and ECDFs obtained by the four solvers. Values in Table~\ref{tab:std} are the average across the tested instances of each benchmark track.
\begin{table}[H]
    \centering
    \begin{tabular}{rcc}
    \toprule
     & score & ECDF \\
    \midrule
 MSE23-w & 0.057 & 0.108 \\
MSE23-uw & 0.096 & 0.112 \\
MSE22-w & 0.054 & 0.118 \\
MSE22-uw & 0.091 & 0.113 \\
\bottomrule
    \end{tabular}
    \caption{The standard deviations of scores and ECDFs}
    \label{tab:std}
\end{table}

\section{B. Results on Groups of Instances}

\begin{figure*}[htb]
    \centering
    \includegraphics[width=0.65\linewidth]{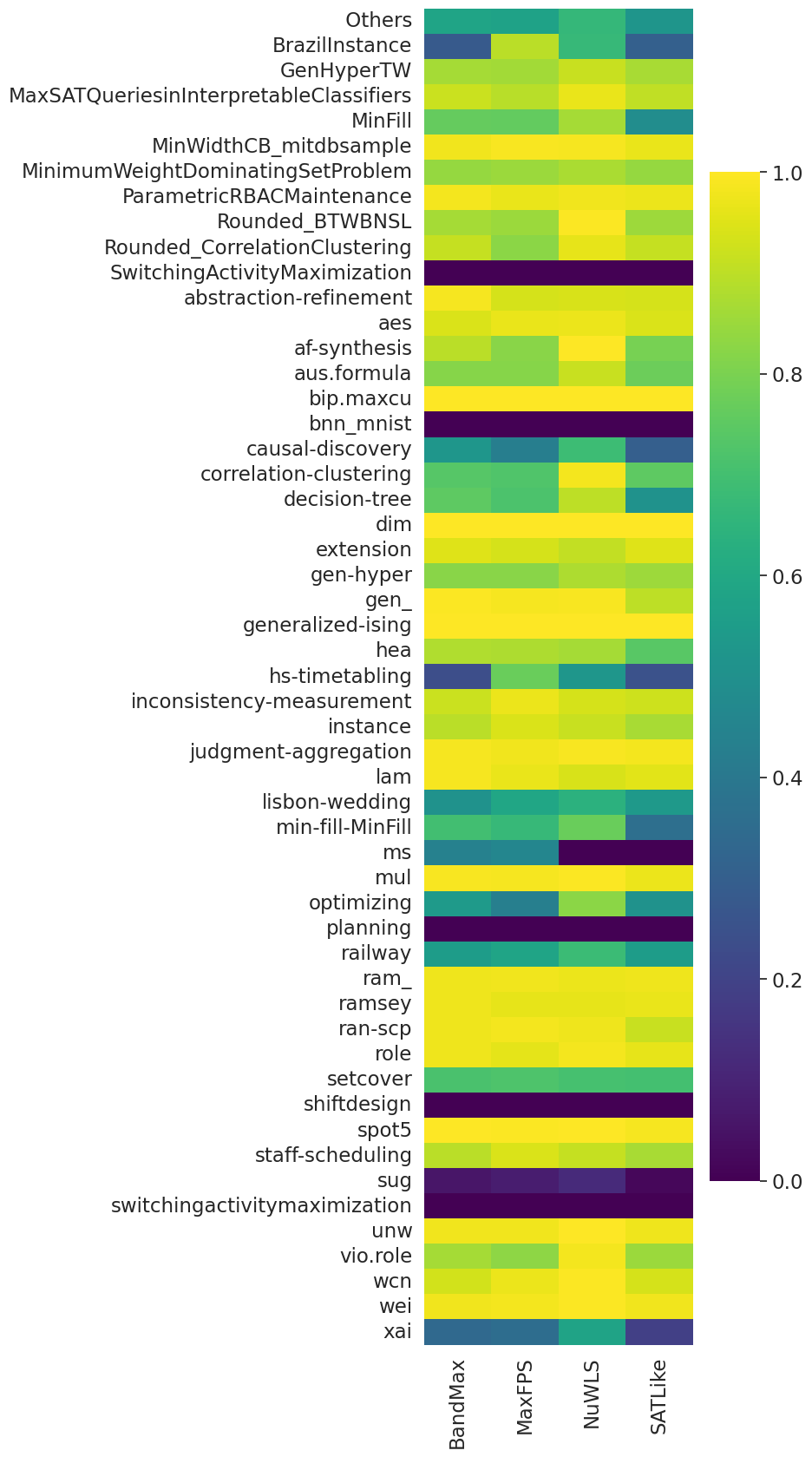}
    \caption{Heatmap of scores of four solvers on groups of instances.}
    \label{fig:app-score}
\end{figure*}

\begin{figure*}[htb]
    \centering
    \includegraphics[width=0.65\linewidth]{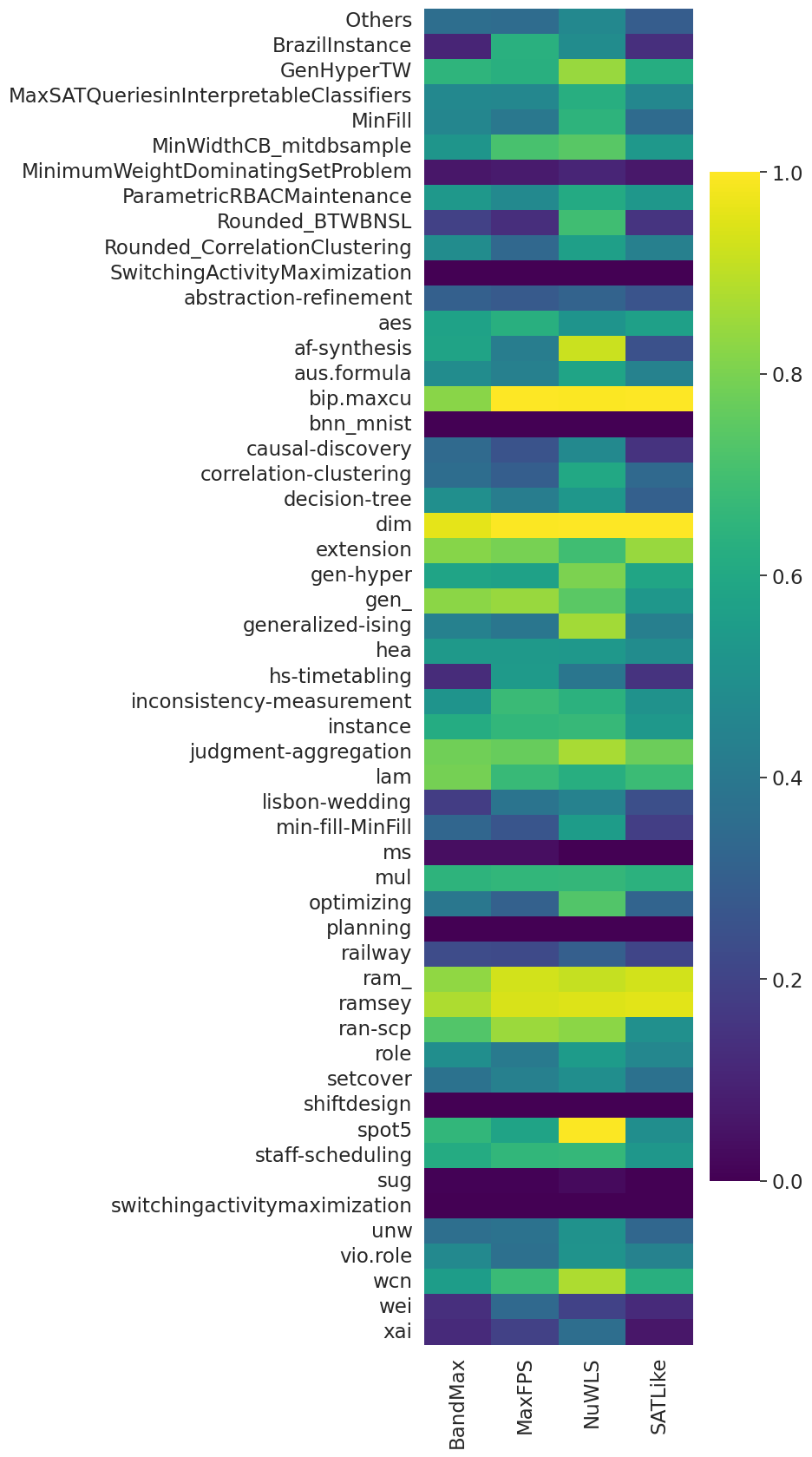}
    \caption{Heatmap of ECDFs of four solvers on groups of instances.}
    \label{fig:app-ecdf}
\end{figure*}

While the scores and ECDFs of hundreds of instances are plotted by colors in Figure~1 and Figure~4 in the submission, we provide the exact average score and ECDF values and their ranks for groups of instances in Table~\ref{tab:app-score}-\ref{tab:app-rank-ecdf}. The groups whose number of instances is less than five are reported together in the group ``Others''. In addition, we provide similar heatmap figures in Figures~\ref{fig:app-score}-\ref{fig:app-ecdf}.

\begin{table*}[ht]
\centering
    \begin{tabular}{rcccc}
\toprule
 & BandMax & MaxFPS & NuWLS & SATLike \\
\midrule
 Others & 0.583 & 0.577 & 0.666 & 0.521 \\
BrazilInstance & 0.279 & 0.895 & 0.669 & 0.302 \\
GenHyperTW & 0.865 & 0.863 & 0.916 & 0.867 \\
MaxSATQueriesinInterpretableClassifiers & 0.919 & 0.892 & 0.967 & 0.904 \\
MinFill & 0.764 & 0.761 & 0.864 & 0.487 \\
MinWidthCB\_mitdbsample & 0.980 & 0.988 & 0.988 & 0.965 \\
MinimumWeightDominatingSetProblem & 0.843 & 0.848 & 0.873 & 0.842 \\
ParametricRBACMaintenance & 0.982 & 0.967 & 0.978 & 0.972 \\
Rounded\_BTWBNSL & 0.865 & 0.849 & 0.995 & 0.854 \\
Rounded\_CorrelationClustering & 0.912 & 0.828 & 0.964 & 0.914 \\
SwitchingActivityMaximization & 0.000 & 0.000 & 0.000 & 0.000 \\
abstraction-refinement & 0.986 & 0.936 & 0.938 & 0.935 \\
aes & 0.943 & 0.965 & 0.971 & 0.943 \\
af-synthesis & 0.895 & 0.822 & 0.999 & 0.796 \\
aus.formula & 0.818 & 0.819 & 0.917 & 0.777 \\
bip.maxcu & 1.000 & 1.000 & 1.000 & 1.000 \\
bnn\_mnist & 0.000 & 0.000 & 0.000 & 0.000 \\
causal-discovery & 0.526 & 0.423 & 0.686 & 0.297 \\
correlation-clustering & 0.736 & 0.726 & 0.981 & 0.752 \\
decision-tree & 0.753 & 0.715 & 0.902 & 0.512 \\
dim & 1.000 & 1.000 & 1.000 & 1.000 \\
extension & 0.953 & 0.937 & 0.908 & 0.952 \\
gen-hyper & 0.820 & 0.822 & 0.877 & 0.852 \\
gen\_ & 0.994 & 0.987 & 0.992 & 0.902 \\
generalized-ising & 0.997 & 0.996 & 1.000 & 0.997 \\
hea & 0.880 & 0.879 & 0.866 & 0.738 \\
hs-timetabling & 0.235 & 0.770 & 0.524 & 0.252 \\
inconsistency-measurement & 0.919 & 0.970 & 0.936 & 0.923 \\
instance & 0.898 & 0.945 & 0.914 & 0.868 \\
judgment-aggregation & 0.984 & 0.978 & 0.988 & 0.982 \\
lam & 0.986 & 0.965 & 0.940 & 0.956 \\
lisbon-wedding & 0.506 & 0.592 & 0.637 & 0.537 \\
min-fill-MinFill & 0.699 & 0.670 & 0.772 & 0.363 \\
ms & 0.436 & 0.457 & 0.000 & 0.000 \\
mul & 0.991 & 0.986 & 0.995 & 0.972 \\
optimizing & 0.539 & 0.426 & 0.825 & 0.508 \\
planning & 0.000 & 0.000 & 0.000 & 0.000 \\
railway & 0.548 & 0.585 & 0.681 & 0.548 \\
ram\_ & 0.974 & 0.978 & 0.971 & 0.975 \\
ramsey & 0.973 & 0.964 & 0.965 & 0.965 \\
ran-scp & 0.975 & 0.981 & 0.975 & 0.915 \\
role & 0.976 & 0.958 & 0.982 & 0.962 \\
setcover & 0.712 & 0.719 & 0.706 & 0.703 \\
shiftdesign & 0.000 & 0.000 & 0.000 & 0.000 \\
spot5 & 0.998 & 0.993 & 1.000 & 0.985 \\
staff-scheduling & 0.897 & 0.945 & 0.914 & 0.868 \\
sug & 0.058 & 0.080 & 0.118 & 0.016 \\
switchingactivitymaximization & 0.000 & 0.000 & 0.000 & 0.000 \\
unw & 0.977 & 0.977 & 0.998 & 0.977 \\
vio.role & 0.863 & 0.830 & 0.984 & 0.849 \\
wcn & 0.933 & 0.970 & 0.996 & 0.937 \\
wei & 0.980 & 0.983 & 0.996 & 0.980 \\
xai & 0.339 & 0.352 & 0.581 & 0.191 \\
\bottomrule
\end{tabular}
\caption{Scores of four solvers on groups of instances}
\label{tab:app-score}
\end{table*}

\begin{table*}[ht]
    \centering
   \begin{tabular}{rcccc}
\toprule
 & BandMax & MaxFPS & NuWLs & SATLike \\
\midrule
 Others & 2.0 & 3.0 & 1.0 & 4.0 \\
BrazilInstance & 4.0 & 1.0 & 2.0 & 3.0 \\
GenHyperTW & 3.0 & 4.0 & 1.0 & 2.0 \\
MaxSATQueriesinInterpretableClassifiers & 2.0 & 4.0 & 1.0 & 3.0 \\
MinFill & 2.0 & 3.0 & 1.0 & 4.0 \\
MinWidthCB\_mitdbsample & 3.0 & 1.0 & 2.0 & 4.0 \\
MinimumWeightDominatingSetProblem & 3.0 & 2.0 & 1.0 & 4.0 \\
ParametricRBACMaintenance & 1.0 & 4.0 & 2.0 & 3.0 \\
Rounded\_BTWBNSL & 2.0 & 4.0 & 1.0 & 3.0 \\
Rounded\_CorrelationClustering & 3.0 & 4.0 & 1.0 & 2.0 \\
SwitchingActivityMaximization & 2.5 & 2.5 & 2.5 & 2.5 \\
abstraction-refinement & 1.0 & 3.0 & 2.0 & 4.0 \\
aes & 4.0 & 2.0 & 1.0 & 3.0 \\
af-synthesis & 2.0 & 3.0 & 1.0 & 4.0 \\
aus.formula & 3.0 & 2.0 & 1.0 & 4.0 \\
bip.maxcu & 2.5 & 2.5 & 2.5 & 2.5 \\
bnn\_mnist & 2.5 & 2.5 & 2.5 & 2.5 \\
causal-discovery & 2.0 & 3.0 & 1.0 & 4.0 \\
correlation-clustering & 3.0 & 4.0 & 1.0 & 2.0 \\
decision-tree & 2.0 & 3.0 & 1.0 & 4.0 \\
dim & 2.5 & 2.5 & 2.5 & 2.5 \\
extension & 1.0 & 3.0 & 4.0 & 2.0 \\
gen-hyper & 4.0 & 3.0 & 1.0 & 2.0 \\
gen\_ & 1.0 & 3.0 & 2.0 & 4.0 \\
generalized-ising & 3.0 & 4.0 & 1.0 & 2.0 \\
hea & 1.0 & 2.0 & 3.0 & 4.0 \\
hs-timetabling & 4.0 & 1.0 & 2.0 & 3.0 \\
inconsistency-measurement & 4.0 & 1.0 & 2.0 & 3.0 \\
instance & 3.0 & 1.0 & 2.0 & 4.0 \\
judgment-aggregation & 2.0 & 4.0 & 1.0 & 3.0 \\
lam & 1.0 & 2.0 & 4.0 & 3.0 \\
lisbon-wedding & 4.0 & 2.0 & 1.0 & 3.0 \\
min-fill-MinFill & 2.0 & 3.0 & 1.0 & 4.0 \\
ms & 2.0 & 1.0 & 3.5 & 3.5 \\
mul & 2.0 & 3.0 & 1.0 & 4.0 \\
optimizing & 2.0 & 4.0 & 1.0 & 3.0 \\
planning & 2.5 & 2.5 & 2.5 & 2.5 \\
railway & 3.0 & 2.0 & 1.0 & 4.0 \\
ram\_ & 3.0 & 1.0 & 4.0 & 2.0 \\
ramsey & 1.0 & 4.0 & 3.0 & 2.0 \\
ran-scp & 2.0 & 1.0 & 3.0 & 4.0 \\
role & 2.0 & 4.0 & 1.0 & 3.0 \\
setcover & 2.0 & 1.0 & 3.0 & 4.0 \\
shiftdesign & 2.5 & 2.5 & 2.5 & 2.5 \\
spot5 & 2.0 & 3.0 & 1.0 & 4.0 \\
staff-scheduling & 3.0 & 1.0 & 2.0 & 4.0 \\
sug & 3.0 & 2.0 & 1.0 & 4.0 \\
switchingactivitymaximization & 2.5 & 2.5 & 2.5 & 2.5 \\
unw & 3.0 & 2.0 & 1.0 & 4.0 \\
vio.role & 2.0 & 4.0 & 1.0 & 3.0 \\
wcn & 4.0 & 2.0 & 1.0 & 3.0 \\
wei & 4.0 & 2.0 & 1.0 & 3.0 \\
xai & 3.0 & 2.0 & 1.0 & 4.0 \\
\bottomrule
\end{tabular}
    \caption{The rank of four solvers regarding scores on groups of instances}
    \label{tab:app-rank-score}
\end{table*}

\begin{table*}[htb!]
    \centering
    \begin{tabular}{rcccc}
\toprule
 & BandMax & MaxFPS & NuWLS & SATLike \\
\midrule
 Others & 0.359 & 0.350 & 0.467 & 0.293 \\
BrazilInstance & 0.105 & 0.635 & 0.481 & 0.135 \\
GenHyperTW & 0.651 & 0.633 & 0.846 & 0.619 \\
MaxSATQueriesinInterpretableClassifiers & 0.466 & 0.463 & 0.626 & 0.462 \\
MinFill & 0.458 & 0.400 & 0.646 & 0.347 \\
MinWidthCB\_mitdbsample & 0.521 & 0.711 & 0.739 & 0.532 \\
MinimumWeightDominatingSetProblem & 0.059 & 0.071 & 0.102 & 0.066 \\
ParametricRBACMaintenance & 0.534 & 0.470 & 0.607 & 0.530 \\
Rounded\_BTWBNSL & 0.194 & 0.132 & 0.694 & 0.149 \\
Rounded\_CorrelationClustering & 0.481 & 0.334 & 0.561 & 0.433 \\
SwitchingActivityMaximization & 0.000 & 0.000 & 0.000 & 0.000 \\
abstraction-refinement & 0.301 & 0.278 & 0.315 & 0.260 \\
aes & 0.577 & 0.630 & 0.517 & 0.563 \\
af-synthesis & 0.579 & 0.418 & 0.921 & 0.250 \\
aus.formula & 0.483 & 0.432 & 0.583 & 0.443 \\
bip.maxcu & 0.824 & 1.000 & 0.995 & 1.000 \\
bnn\_mnist & 0.000 & 0.000 & 0.000 & 0.000 \\
causal-discovery & 0.344 & 0.256 & 0.472 & 0.146 \\
correlation-clustering & 0.352 & 0.297 & 0.601 & 0.337 \\
decision-tree & 0.493 & 0.420 & 0.528 & 0.302 \\
dim & 0.961 & 0.995 & 1.000 & 0.998 \\
extension & 0.818 & 0.795 & 0.692 & 0.847 \\
gen-hyper & 0.584 & 0.574 & 0.802 & 0.586 \\
gen\_ & 0.826 & 0.846 & 0.745 & 0.531 \\
generalized-ising & 0.435 & 0.394 & 0.861 & 0.426 \\
hea & 0.539 & 0.536 & 0.533 & 0.481 \\
hs-timetabling & 0.121 & 0.539 & 0.392 & 0.147 \\
inconsistency-measurement & 0.517 & 0.679 & 0.638 & 0.507 \\
instance & 0.617 & 0.659 & 0.669 & 0.533 \\
judgment-aggregation & 0.784 & 0.766 & 0.868 & 0.774 \\
lam & 0.792 & 0.673 & 0.628 & 0.680 \\
lisbon-wedding & 0.177 & 0.385 & 0.442 & 0.239 \\
min-fill-MinFill & 0.324 & 0.263 & 0.549 & 0.182 \\
ms & 0.038 & 0.037 & 0.000 & 0.000 \\
mul & 0.645 & 0.658 & 0.662 & 0.638 \\
optimizing & 0.398 & 0.308 & 0.729 & 0.320 \\
planning & 0.000 & 0.000 & 0.000 & 0.000 \\
railway & 0.230 & 0.226 & 0.300 & 0.204 \\
ram\_ & 0.833 & 0.932 & 0.911 & 0.931 \\
ramsey & 0.876 & 0.941 & 0.947 & 0.955 \\
ran-scp & 0.728 & 0.849 & 0.826 & 0.498 \\
role & 0.492 & 0.407 & 0.545 & 0.462 \\
setcover & 0.377 & 0.433 & 0.488 & 0.371 \\
shiftdesign & 0.000 & 0.000 & 0.000 & 0.000 \\
spot5 & 0.658 & 0.581 & 0.993 & 0.489 \\
staff-scheduling & 0.611 & 0.659 & 0.667 & 0.529 \\
sug & 0.006 & 0.005 & 0.024 & 0.003 \\
switchingactivitymaximization & 0.000 & 0.000 & 0.000 & 0.000 \\
unw & 0.361 & 0.378 & 0.508 & 0.330 \\
vio.role & 0.470 & 0.365 & 0.515 & 0.442 \\
wcn & 0.551 & 0.678 & 0.876 & 0.631 \\
wei & 0.135 & 0.338 & 0.196 & 0.120 \\
xai & 0.120 & 0.195 & 0.358 & 0.060 \\
\bottomrule
\end{tabular}

    \caption{ECDFs of four solvers on groups of instances}
    \label{tab:app-ecdf}
\end{table*}

\begin{table*}[htb!]
\centering
   \begin{tabular}{rcccc}
\toprule
& BandMax & MaxFPS & NuWLS & SATLike \\
\midrule
 Others & 2.0 & 3.0 & 1.0 & 4.0 \\
BrazilInstance & 4.0 & 1.0 & 2.0 & 3.0 \\
GenHyperTW & 2.0 & 3.0 & 1.0 & 4.0 \\
MaxSATQueriesinInterpretableClassifiers & 2.0 & 3.0 & 1.0 & 4.0 \\
MinFill & 2.0 & 3.0 & 1.0 & 4.0 \\
MinWidthCB\_mitdbsample & 4.0 & 2.0 & 1.0 & 3.0 \\
MinimumWeightDominatingSetProblem & 4.0 & 2.0 & 1.0 & 3.0 \\
ParametricRBACMaintenance & 2.0 & 4.0 & 1.0 & 3.0 \\
Rounded\_BTWBNSL & 2.0 & 4.0 & 1.0 & 3.0 \\
Rounded\_CorrelationClustering & 2.0 & 4.0 & 1.0 & 3.0 \\
SwitchingActivityMaximization & 2.5 & 2.5 & 2.5 & 2.5 \\
abstraction-refinement & 2.0 & 3.0 & 1.0 & 4.0 \\
aes & 2.0 & 1.0 & 4.0 & 3.0 \\
af-synthesis & 2.0 & 3.0 & 1.0 & 4.0 \\
aus.formula & 2.0 & 4.0 & 1.0 & 3.0 \\
bip.maxcu & 4.0 & 1.5 & 3.0 & 1.5 \\
bnn\_mnist & 2.5 & 2.5 & 2.5 & 2.5 \\
causal-discovery & 2.0 & 3.0 & 1.0 & 4.0 \\
correlation-clustering & 2.0 & 4.0 & 1.0 & 3.0 \\
decision-tree & 2.0 & 3.0 & 1.0 & 4.0 \\
dim & 4.0 & 3.0 & 1.0 & 2.0 \\
extension & 2.0 & 3.0 & 4.0 & 1.0 \\
gen-hyper & 3.0 & 4.0 & 1.0 & 2.0 \\
gen\_ & 2.0 & 1.0 & 3.0 & 4.0 \\
generalized-ising & 2.0 & 4.0 & 1.0 & 3.0 \\
hea & 1.0 & 2.0 & 3.0 & 4.0 \\
hs-timetabling & 4.0 & 1.0 & 2.0 & 3.0 \\
inconsistency-measurement & 3.0 & 1.0 & 2.0 & 4.0 \\
instance & 3.0 & 2.0 & 1.0 & 4.0 \\
judgment-aggregation & 2.0 & 4.0 & 1.0 & 3.0 \\
lam & 1.0 & 3.0 & 4.0 & 2.0 \\
lisbon-wedding & 4.0 & 2.0 & 1.0 & 3.0 \\
min-fill-MinFill & 2.0 & 3.0 & 1.0 & 4.0 \\
ms & 1.0 & 2.0 & 3.5 & 3.5 \\
mul & 3.0 & 2.0 & 1.0 & 4.0 \\
optimizing & 2.0 & 4.0 & 1.0 & 3.0 \\
planning & 2.5 & 2.5 & 2.5 & 2.5 \\
railway & 2.0 & 3.0 & 1.0 & 4.0 \\
ram\_ & 4.0 & 1.0 & 3.0 & 2.0 \\
ramsey & 4.0 & 3.0 & 2.0 & 1.0 \\
ran-scp & 3.0 & 1.0 & 2.0 & 4.0 \\
role & 2.0 & 4.0 & 1.0 & 3.0 \\
setcover & 3.0 & 2.0 & 1.0 & 4.0 \\
shiftdesign & 2.5 & 2.5 & 2.5 & 2.5 \\
spot5 & 2.0 & 3.0 & 1.0 & 4.0 \\
staff-scheduling & 3.0 & 2.0 & 1.0 & 4.0 \\
sug & 2.0 & 3.0 & 1.0 & 4.0 \\
switchingactivitymaximization & 2.5 & 2.5 & 2.5 & 2.5 \\
unw & 3.0 & 2.0 & 1.0 & 4.0 \\
vio.role & 2.0 & 4.0 & 1.0 & 3.0 \\
wcn & 4.0 & 2.0 & 1.0 & 3.0 \\
wei & 3.0 & 1.0 & 2.0 & 4.0 \\
xai & 3.0 & 2.0 & 1.0 & 4.0 \\
\bottomrule
\end{tabular}

    \caption{The rank of of four solvers regarding ECDFs on groups of instances}
    \label{tab:app-rank-ecdf}
\end{table*}

\section{C. Detailed Results of Configurations obtained by SMAC}

Due to the limited space, we can provide only the average results of five configurations for each tuning scenario in Table 4 in the submission. Here, we provide in Tables~\ref{tab:5-score-w}-\ref{tab:5-ecdf-uw} the result of configurations obtained by each SMAC run. In addition, we have calculated the Wilcoxon signed-rank test, comparing the results of using ECDF as the cost function with the others. P-values are reported in the tables.

\begin{table*}[htb]
    \centering
{    
    \begin{tabular}{rccc|ccc|ccc|ccc}
\hline
 & \multicolumn{3}{c}{BandMax} & \multicolumn{3}{c}{MaxFPS} & \multicolumn{3}{c}{NuWLS} & \multicolumn{3}{c}{SATLike} \\
\midrule
 & Best-f & Norm-f &  ECDF & Best-f &  Norm-f & ECDF & Best-f  &  Norm-f & ECDF & Best-f &  Norm-f & ECDF \\
\midrule
run1 &	0.857 	&	0.850 	&	0.858 	&	0.848 	&	0.852 	&	0.847 	&	0.799 	&	0.808 	&	0.806 	&	0.795 	&	0.560 	&	0.794 \\
run2 &	0.858 	&	0.850 	&	0.859 	&	0.845 	&	0.845 	&	0.847 	&	0.787 	&	0.782 	&	0.820 	&	0.771 	&	0.562 	&	0.842 \\
run3 &	0.853 	&	0.851 	&	0.856 	&	0.842 	&	0.850 	&	0.847 	&	0.786 	&	0.808 	&	0.809 	&	0.498 	&	0.470 	&	0.656 \\
run4 &	0.858 	&	0.850 	&	0.857 	&	0.843 	&	0.847 	&	0.846 	&	0.794 	&	0.808 	&	0.814 	&	0.478 	&	0.476 	&	0.656 \\
run5 &	0.858 	&	0.848 	&	0.858 	&	0.841 	&	0.855 	&	0.853 	&	0.811 	&	0.837 	&	0.839 	&	0.778 	&	0.753 	&	0.841 \\
\midrule
Avg. &	0.857	&	0.850	&	\textbf{0.858}	&	0.844	&	\textbf{0.850}	&	0.848	&	0.795	&	0.809	&	\textbf{0.818}	&	0.664	&	0.564	&	\textbf{0.758} \\
Best &	0.858	&	0.851	&	\textbf{0.859}	&	0.848	&	\textbf{0.855}	&	0.853	&	0.811	&	0.837	&	\textbf{0.839}	&	0.795	&	0.753	&	\textbf{0.842} \\
pvalue	&	0.313	&	0.063	&	-	&	0.125	&	0.313	&	-	&	0.063	&	0.125	&	-	&	0.125	&	0.063	&	-\\
\bottomrule
\end{tabular}

}
\caption{Scores of five configurations obtained by each SMAC setting on MSE23-w. Larger values indicate better performances.}
\label{tab:5-score-w}
\end{table*}

\begin{table*}[htb]
    \centering
{    
    \begin{tabular}{rccc|ccc|ccc|ccc}
\hline
 & \multicolumn{3}{c}{BandMax} & \multicolumn{3}{c}{MaxFPS} & \multicolumn{3}{c}{NuWLS} & \multicolumn{3}{c}{SATLike} \\
\midrule
 & Best-f & Norm-f &  ECDF & Best-f &  Norm-f & ECDF & Best-f  &  Norm-f & ECDF & Best-f &  Norm-f & ECDF \\
\midrule
run1&	0.816 	&	0.803 	&	0.814 	&	0.788 	&	0.771 	&	0.767 	&	0.839 	&	0.817 	&	0.839 	&	0.711 	&	0.755 	&	0.855 \\
run2&	0.819 	&	0.802 	&	0.818 	&	0.778 	&	0.765 	&	0.768 	&	0.806 	&	0.838 	&	0.851 	&	0.732 	&	0.595 	&	0.805 \\
run3&	0.812 	&	0.786 	&	0.810 	&	0.770 	&	0.782 	&	0.770 	&	0.810 	&	0.830 	&	0.888 	&	0.553 	&	0.756 	&	0.622 \\
run4&	0.815 	&	0.801 	&	0.815 	&	0.782 	&	0.773 	&	0.780 	&	0.844 	&	0.835 	&	0.836 	&	0.539 	&	0.723 	&	0.824 \\
run5&	0.814 	&	0.802 	&	0.814 	&	0.763 	&	0.775 	&	0.762 	&	0.821 	&	0.815 	&	0.818 	&	0.549 	&	0.754 	&	0.846 \\
\midrule
Avg.&	\textbf{0.815}	&	0.799	&	0.814	&	\textbf{0.776}	&	0.773	&	0.769	&	0.824	&	0.827	&	\textbf{0.846}	&	0.617	&	0.717	&	\textbf{0.790} \\
Best&	\textbf{0.819}	&	0.803	&	0.818	&	\textbf{0.788}	&	0.782	&	0.780	&	0.844	&	0.838	&	\textbf{0.888}	&	0.732	&	0.756	&	\textbf{0.855} \\
pvalue	&	0.313	&	0.063	&	-	&	0.063	&	0.438	&	-	&	0.625	&	0.063	&	-	&	0.063	&	0.438	&	-\\
\bottomrule
\end{tabular}

}
\caption{Scores of five configurations obtained by each SMAC setting on MSE23-uw. Larger values indicate better performances.}
\label{tab:5-score-uw}
\end{table*}

\begin{table*}[htb]
    \centering
{    
    \begin{tabular}{rccc|ccc|ccc|ccc}
\hline
 & \multicolumn{3}{c}{BandMax} & \multicolumn{3}{c}{MaxFPS} & \multicolumn{3}{c}{NuWLS} & \multicolumn{3}{c}{SATLike} \\
\midrule
 & Best-f & Norm-f &  ECDF & Best-f &  Norm-f & ECDF & Best-f  &  Norm-f & ECDF & Best-f &  Norm-f & ECDF \\
\midrule
run1&	0.426 	&	0.408 	&	0.425 	&	0.429 	&	0.442 	&	0.436 	&	0.437 	&	0.467 	&	0.464 	&	0.359 	&	0.241 	&	0.359 \\
run2&	0.430 	&	0.405 	&	0.430 	&	0.438 	&	0.408 	&	0.437 	&	0.416 	&	0.362 	&	0.473 	&	0.357 	&	0.239 	&	0.435 \\
run3&	0.413 	&	0.406 	&	0.412 	&	0.419 	&	0.421 	&	0.436 	&	0.403 	&	0.467 	&	0.470 	&	0.205 	&	0.172 	&	0.263 \\
run4&	0.424 	&	0.400 	&	0.424 	&	0.410 	&	0.430 	&	0.431 	&	0.427 	&	0.467 	&	0.479 	&	0.160 	&	0.174 	&	0.269 \\
run5&	0.424 	&	0.407 	&	0.422 	&	0.423 	&	0.416 	&	0.412 	&	0.472 	&	0.487 	&	0.511 	&	0.360 	&	0.301 	&	0.439 \\
\midrule
Avg.&	\textbf{0.423} 	&	0.405 	&	\textbf{0.423} 	&	0.424 	&	0.423 	&	\textbf{0.430} 	&	0.431 	&	0.450 	&	\textbf{0.479} 	&	0.288 	&	0.225 	&	\textbf{0.353} \\
Best&	\textbf{0.430} 	&	0.408 	&	\textbf{0.430}	&	0.438 	&	\textbf{0.442} 	&	0.437 	&	0.472 	&	0.487 	&	\textbf{0.511} 	&	0.360 	&	0.301 	&	\textbf{0.439} \\
pvalue	&	0.063	&	0.063	&	-	&	0.438	&	0.625	&	-	&	0.063	&	0.188	&	-	&	0.063	&	0.063	&	- \\

\bottomrule
\end{tabular}

}
\caption{ECDFs of five configurations obtained by each SMAC setting on MSE23-w. Larger values indicate better performances.}
\label{tab:5-ecdf-w}
\end{table*}

\begin{table*}[htb]
    \centering
{    
    \begin{tabular}{rccc|ccc|ccc|ccc}
\hline
 & \multicolumn{3}{c}{BandMax} & \multicolumn{3}{c}{MaxFPS} & \multicolumn{3}{c}{NuWLS} & \multicolumn{3}{c}{SATLike} \\
\midrule
 & Best-f & Norm-f &  ECDF & Best-f &  Norm-f & ECDF & Best-f  &  Norm-f & ECDF & Best-f &  Norm-f & ECDF \\
\midrule
run1&	0.551 	&	0.538 	&	0.551 	&	0.526 	&	0.499 	&	0.492 	&	0.522 	&	0.476 	&	0.529 	&	0.409 	&	0.423 	&	0.510 \\
run2&	0.558 	&	0.538 	&	0.557 	&	0.516 	&	0.480 	&	0.490 	&	0.482 	&	0.526 	&	0.552 	&	0.455 	&	0.397 	&	0.542 \\
run3&	0.537 	&	0.510 	&	0.534 	&	0.493 	&	0.523 	&	0.496 	&	0.450 	&	0.500 	&	0.604 	&	0.345 	&	0.422 	&	0.372 \\
run4&	0.549 	&	0.538 	&	0.548 	&	0.510 	&	0.508 	&	0.515 	&	0.509 	&	0.487 	&	0.498 	&	0.315 	&	0.454 	&	0.477 \\
run5&	0.552 	&	0.538 	&	0.549 	&	0.480 	&	0.503 	&	0.488 	&	0.476 	&	0.468 	&	0.459 	&	0.341 	&	0.421 	&	0.485 \\
\midrule
Avg.&	\textbf{0.549} 	&	0.532 	&	0.548 	&	\textbf{0.505} 	&	0.503 	&	0.496 	&	0.488 	&	0.491 	&	\textbf{0.528} 	&	0.373 	&	0.423 	&	\textbf{0.477} \\
Best&	\textbf{0.558} 	&	0.538 	&	0.557 	&	\textbf{0.526} 	&	0.523 	&	0.515 	&	0.522 	&	0.526 	&	\textbf{0.604} 	&	0.455 	&	0.454 	&	\textbf{0.542} \\
pvalue	&	0.063	&	0.063	&	-	&	0.813	&	0.625	&	-	&	0.625	&	0.125	&	-	&	0.063	&	0.188	&	- \\

\bottomrule
\end{tabular}

}
\caption{ECDFs of five configurations obtained by each SMAC setting on MSE23-uw. Larger values indicate better performances.}
\label{tab:5-ecdf-uw}
\end{table*}

\section{D. Results of Using Cost Functions Normalized across Multiple Instances}
\begin{figure}[htb!]
    \centering
    \subcaptionbox{BandMax}{\includegraphics[trim=5 5 0 5, clip,width=\linewidth]{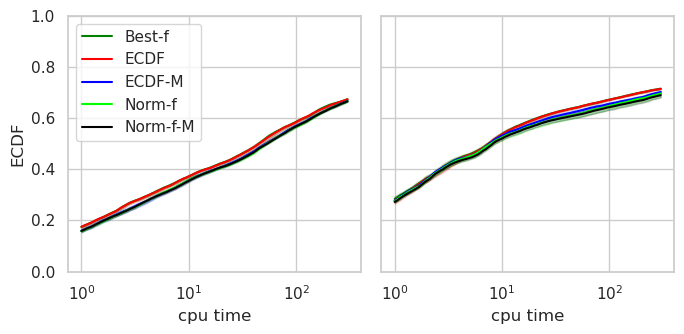}}
     \subcaptionbox{MaxFPS}{\includegraphics[trim=5 5 0 5, clip,width=\linewidth]{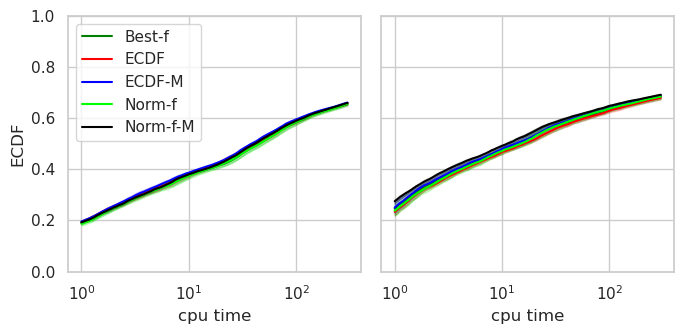}}
    \subcaptionbox{NuWLS}{\includegraphics[trim=5 5 0 5, clip,width=\linewidth]{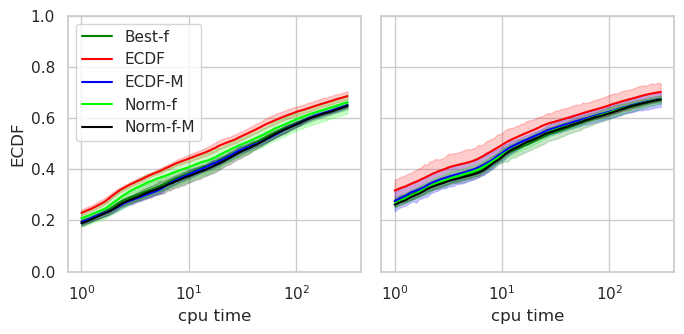}}
     \subcaptionbox{SATLike}{\includegraphics[trim=5 5 0 5, clip,width=\linewidth]{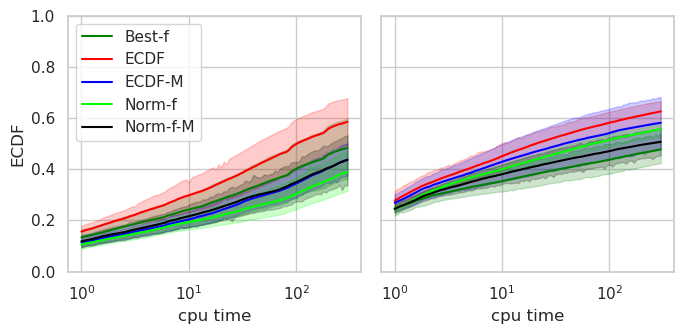}}
    \caption{ECDFs of the configurations obtained by tuning for Best-f, Norm-f, Norm-f-M, ECDF, and ECDF-M for \textbf{Left:} MSE23-w and \textbf{Right:} MSE23-uw. Results plotted in one line are from five configurations obtained by independent runs of SMAC.}
    \label{fig:tunedecdf-m}
\end{figure}

As mentioned in the submission, we have tested using the average of ECDF and Norm-f values across 20 instances, namely \textbf{ECDF-M} and \textbf{Norm-f-M}, as the cost function because the MaxSAT Evaluation competitions measure solvers across many instances. We provide in this appendix the corresponding scores and ECDFs in Tables~\ref{tab:5-score-mw}-\ref{tab:5-ecdf-muw}. Also, we plot the ECDF lines along cpu time in Figure~\ref{fig:tunedecdf-m}.

\begin{table*}[htb]
    \centering
{    
    \begin{tabular}{rcc|cc|cc|cc}
\hline
 & \multicolumn{2}{c}{BandMax} & \multicolumn{2}{c}{MaxFPS} & \multicolumn{2}{c}{NuWLS} & \multicolumn{2}{c}{SATLike} \\
\midrule
&	ECDF-M	&	Norm-f-M	&	ECDF-M	&	Norm-f-M	&	ECDF-M	&	Norm-f-M	&	ECDF-mul	&	Norm-f-M \\
\midrule
run1&	0.849 	&	0.851 	&	0.845 	&	0.842 	&	0.804 	&	0.793 	&	0.727 	&	0.778 \\
run2&	0.854 	&	0.851 	&	0.849 	&	0.849 	&	0.821 	&	0.810 	&	0.696 	&	0.515 \\
run3&	0.856 	&	0.855 	&	0.846 	&	0.845 	&	0.823 	&	0.807 	&	0.487 	&	0.572 \\
run4&	0.853 	&	0.856 	&	0.848 	&	0.855 	&	0.809 	&	0.798 	&	0.575 	&	0.728 \\
run5&	0.851 	&	0.848 	&	0.847 	&	0.848 	&	0.816 	&	0.789 	&	0.562 	&	0.467 \\
\midrule
Avg.&	0.853	&	0.852	&	0.847	&	0.848	&	0.815	&	0.799	&	0.609	&	0.612\\
Best&	0.856	&	0.856	&	0.849	&	0.855	&	0.823	&	0.810	&	0.727	&	0.778\\

\bottomrule
\end{tabular}

}
\caption{Scores of five configurations obtained by each SMAC setting on MSE23-w. Larger values indicate better performances. The results are from using the cost functions normalized across multiple instances.}
\label{tab:5-score-mw}
\end{table*}

\begin{table*}[htb]
    \centering
{    
    \begin{tabular}{rcc|cc|cc|cc}
\hline
 & \multicolumn{2}{c}{BandMax} & \multicolumn{2}{c}{MaxFPS} & \multicolumn{2}{c}{NuWLS} & \multicolumn{2}{c}{SATLike} \\
\midrule
&	ECDF-M	&	Norm-f-M	&	ECDF-M	&	Norm-f-M	&	ECDF-M	&	Norm-f-M	&	ECDF-mul	&	Norm-f-M \\
\midrule
run1	&	0.809 	&	0.796 	&	0.786 	&	0.777 	&	0.826 	&	0.823 	&	0.858 	&	0.575 	\\
run2	&	0.799 	&	0.783 	&	0.774 	&	0.783 	&	0.828 	&	0.825 	&	0.841 	&	0.531 	\\
run3	&	0.808 	&	0.808 	&	0.772 	&	0.778 	&	0.793 	&	0.828 	&	0.772 	&	0.636 	\\
run4	&	0.795 	&	0.800 	&	0.784 	&	0.783 	&	0.833 	&	0.838 	&	0.527 	&	0.726 	\\
run5	&	0.805 	&	0.786 	&	0.770 	&	0.786 	&	0.854 	&	0.834 	&	0.582 	&	0.680 	\\
\midrule
Avg.	&	0.803	&	0.795	&	0.777	&	0.781	&	0.827	&	0.830	&	0.716	&	0.630	\\
Best	&	0.809	&	0.808	&	0.786	&	0.786	&	0.854	&	0.838	&	0.858	&	0.726	\\

\bottomrule
\end{tabular}

}
\caption{Scores of five configurations obtained by each SMAC setting on MSE23-uw. Larger values indicate better performances. The results are from using the cost functions normalized across multiple instances.}
\label{tab:5-score-muw}
\end{table*}

\begin{table*}[htb]
    \centering
{    
    \begin{tabular}{rcc|cc|cc|cc}
\hline
 & \multicolumn{2}{c}{BandMax} & \multicolumn{2}{c}{MaxFPS} & \multicolumn{2}{c}{NuWLS} & \multicolumn{2}{c}{SATLike} \\
\midrule
&	ECDF-M	&	Norm-f-M	&	ECDF-M	&	Norm-f-M	&	ECDF-M	&	Norm-f-M	&	ECDF-mul	&	Norm-f-M \\
\midrule
run1	&	0.404 	&	0.408 	&	0.434 	&	0.429 	&	0.419 	&	0.421 	&	0.308 	&	0.348 	\\
run2	&	0.411 	&	0.402 	&	0.435 	&	0.431 	&	0.413 	&	0.399 	&	0.288 	&	0.230 	\\
run3	&	0.404 	&	0.402 	&	0.431 	&	0.420 	&	0.411 	&	0.451 	&	0.187 	&	0.227 	\\
run4	&	0.413 	&	0.411 	&	0.433 	&	0.432 	&	0.445 	&	0.430 	&	0.224 	&	0.307 	\\
run5	&	0.403 	&	0.403 	&	0.443 	&	0.432 	&	0.439 	&	0.414 	&	0.235 	&	0.168 	\\
\midrule
Avg.	&	0.407	&	0.405	&	0.435	&	0.429	&	0.425	&	0.423	&	0.248	&	0.256	\\
Best	&	0.413	&	0.411	&	0.443	&	0.432	&	0.445	&	0.451	&	0.308	&	0.348	\\

\bottomrule
\end{tabular}

}
\caption{ECDFs of five configurations obtained by each SMAC setting on MSE23-w. Larger values indicate better performances. The results are from using the cost functions normalized across multiple instances.}
\label{tab:5-ecdf-mw}
\end{table*}

\begin{table*}[htb]
    \centering
{    
    \begin{tabular}{rcc|cc|cc|cc}
\hline
 & \multicolumn{2}{c}{BandMax} & \multicolumn{2}{c}{MaxFPS} & \multicolumn{2}{c}{NuWLS} & \multicolumn{2}{c}{SATLike} \\
\midrule
&	ECDF-M	&	Norm-f-M	&	ECDF-M	&	Norm-f-M	&	ECDF-M	&	Norm-f-M	&	ECDF-mul	&	Norm-f-M \\
\midrule
run1	&	0.541 	&	0.528 	&	0.527 	&	0.515 	&	0.519 	&	0.478 	&	0.522 	&	0.368 	\\
run2	&	0.536 	&	0.515 	&	0.499 	&	0.524 	&	0.484 	&	0.495 	&	0.521 	&	0.325 	\\
run3	&	0.542 	&	0.540 	&	0.499 	&	0.517 	&	0.446 	&	0.485 	&	0.518 	&	0.405 	\\
run4	&	0.531 	&	0.534 	&	0.522 	&	0.522 	&	0.501 	&	0.479 	&	0.322 	&	0.433 	\\
run5	&	0.542 	&	0.524 	&	0.497 	&	0.530 	&	0.545 	&	0.498 	&	0.364 	&	0.455 	\\
\midrule
Avg.	&	0.538	&	0.528	&	0.509	&	0.522	&	0.499	&	0.487	&	0.449	&	0.397	\\
Best	&	0.542	&	0.540	&	0.527	&	0.530	&	0.545	&	0.498	&	0.522	&	0.455	\\

\bottomrule
\end{tabular}

}
\caption{ECDFs of five configurations obtained by each SMAC setting on MSE23-uw. Larger values indicate better performances. The results are from using the cost functions normalized across multiple instances.}
\label{tab:5-ecdf-muw}
\end{table*}

